\documentclass{article}

 \usepackage[dblblindworkshop, final]{neurips_2025}
\workshoptitle{Scaling Environments for Agents (SEA)}

\usepackage[utf8]{inputenc} 
\usepackage[T1]{fontenc}    
\usepackage{hyperref}       
\usepackage{url}            
\usepackage{booktabs}       
\usepackage{amsfonts}       
\usepackage{nicefrac}       
\usepackage{microtype}      

\usepackage{inconsolata}
\usepackage{hyperref}
\usepackage{url}
\usepackage{hyperref}       
\usepackage{url}            
\usepackage{booktabs}       
\usepackage{amsfonts}       
\usepackage{nicefrac}       
\usepackage{microtype}      
\usepackage{url}            
\usepackage{booktabs}       
\usepackage{amsfonts}       
\usepackage{nicefrac}       
\usepackage{microtype}      
\usepackage{subcaption}
\usepackage{amsmath,amsfonts,amssymb}
\usepackage{tabularx}
\usepackage{multirow}
\usepackage{graphicx}
\usepackage{color}
\usepackage{tcolorbox}
\usepackage{CJK}
\usepackage{adjustbox}
\usepackage{colortbl}
\usepackage{multicol}
\usepackage{vwcol} 
\newsavebox{\algleft}
\newsavebox{\algright}
\usepackage{bm}
\usepackage{booktabs}
\usepackage{amsmath,stackengine}
\usepackage[linesnumbered,ruled,vlined]{algorithm2e}
\usepackage{times}
\usepackage{latexsym}
\usepackage{bbm}
\usepackage[T1]{fontenc}    
\usepackage{hyperref}       
\usepackage{url}            
\usepackage{booktabs}       
\usepackage{amsfonts}       
\usepackage{nicefrac}       
\usepackage{microtype}      
\usepackage{amssymb}
\usepackage{pifont}
\usepackage{bbm}
\usepackage{breqn}
\usepackage{longtable}
\usepackage{algpseudocode}
\usepackage{enumitem}
\usepackage[table,xcdraw]{xcolor}
\usepackage{microtype}
\usepackage{enumitem}
\usepackage{amsmath}
\usepackage{chngcntr} 
\usepackage{wrapfig}
\usepackage{listings}
\usepackage{arydshln}
\usepackage{amsmath} 
\definecolor{jsonkey}{rgb}{0.6, 0.2, 0.2}
\definecolor{jsonstring}{rgb}{0.2, 0.6, 0.2}

\lstdefinestyle{jsonstyle}{
    basicstyle=\ttfamily\small,
    stringstyle=\color{jsonstring},
    keywordstyle=\color{jsonkey},
    showstringspaces=false,
    breaklines=true,
    frame=none,
    morekeywords={true,false,null} 
}

\lstdefinestyle{mypython}{
    language=Python,
    basicstyle=\ttfamily\scriptsize,
    keywordstyle=\color{blue},
    stringstyle=\color{darkgreen},
    commentstyle=\color{gray},
    showstringspaces=false,
    breaklines=true,
}

\definecolor{BlueBG}{rgb}{0,0.46,0.71}
\newcommand{\ourmodel}{{\sc{MAS-Zero}}\xspace} 
\newcommand{\ourframework}{{\sc{MAS-Zero}}\xspace} 

\definecolor{innerboxcolor}{HTML}{00B2E6} 
\definecolor{outerboxcolor}{HTML}{D7F6FF} 
\definecolor{lightblue}{HTML}{D7F6FF}

\usepackage{tikz}
\usetikzlibrary{decorations.pathmorphing}

\definecolor{upblue}{HTML}{388E3C}
\definecolor{downred}{HTML}{E53935}

\definecolor{darkgreen}{RGB}{0,128,0}  
\definecolor{darkred}{RGB}{139,0,0}  

\newcommand{\downred}[1]{$_{\color{darkred}\downarrow #1}$}
\newcommand{\upgreen}[1]{$_{\color{darkgreen}\uparrow #1}$}
\definecolor{myblue}{RGB}{31,119,180}    
\definecolor{mygreen}{RGB}{44,160,44}    
\definecolor{myorange}{RGB}{255,127,14}  
\definecolor{mypurple}{RGB}{153, 102, 204}  

\newcommand{\MASbox}[2]{\raisebox{0pt}{\colorbox{#1}{#2}}}

\newcommand{\SingleAgent}[1]{\MASbox{cyan!20}{#1}}
\newcommand{\ManualMAS}[1]{\MASbox{purple!20}{#1}}
\newcommand{\ValidationPrune}[1]{\MASbox{blue!10}{#1}}
\newcommand{\ValidationGen}[1]{\MASbox{blue!30}{#1}}
\newcommand{\TrainingMAS}[1]{\MASbox{green!20}{#1}}
\newcommand{\OurMethod}[1]{\MASbox{orange!20}{#1}}

\newcommand{\MyPurple}[1]{\MASbox{mypurple!20}{#1}}
\newcommand{\MyBlue}[1]{\MASbox{myblue!20}{#1}}
\newcommand{\MyOrange}[1]{\MASbox{myorange!20}{#1}}

\newcommand{\stepbox}[2]{%
  \par\noindent\colorbox{#1!18}{%
    \parbox{\dimexpr\linewidth-2\fboxsep}{\textbf{#2}}%
  }\par\smallskip
}

\title{\ourframework: Designing Multi-Agent Systems with Zero Supervision}

\author{%
  Zixuan Ke$^{\ddagger}$, Austin Xu$^{\ddagger}$, Yifei Ming$^{\ddagger}$, Xuan-Phi Nguyen$^{\ddagger}$, Ryan Chin$^{\diamond}$, \\ \textbf{Caiming Xiong}$^{\ddagger}$, \textbf{Shafiq Joty}$^{\ddagger}$\\
  $^{\ddagger}$Salesforce AI Research~~~~~~~$^{\diamond}$Massachusetts Institute of Technology\\
\texttt{\{zixuan.ke,cxiong,sjoty\}@salesforce.com} \\
    \includegraphics[width=0.4cm]{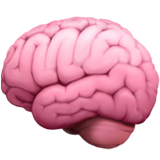} \small Project Page: \url{https://vincent950129.github.io/mas-design/mas_zero} \\
    \includegraphics[width=0.4cm]{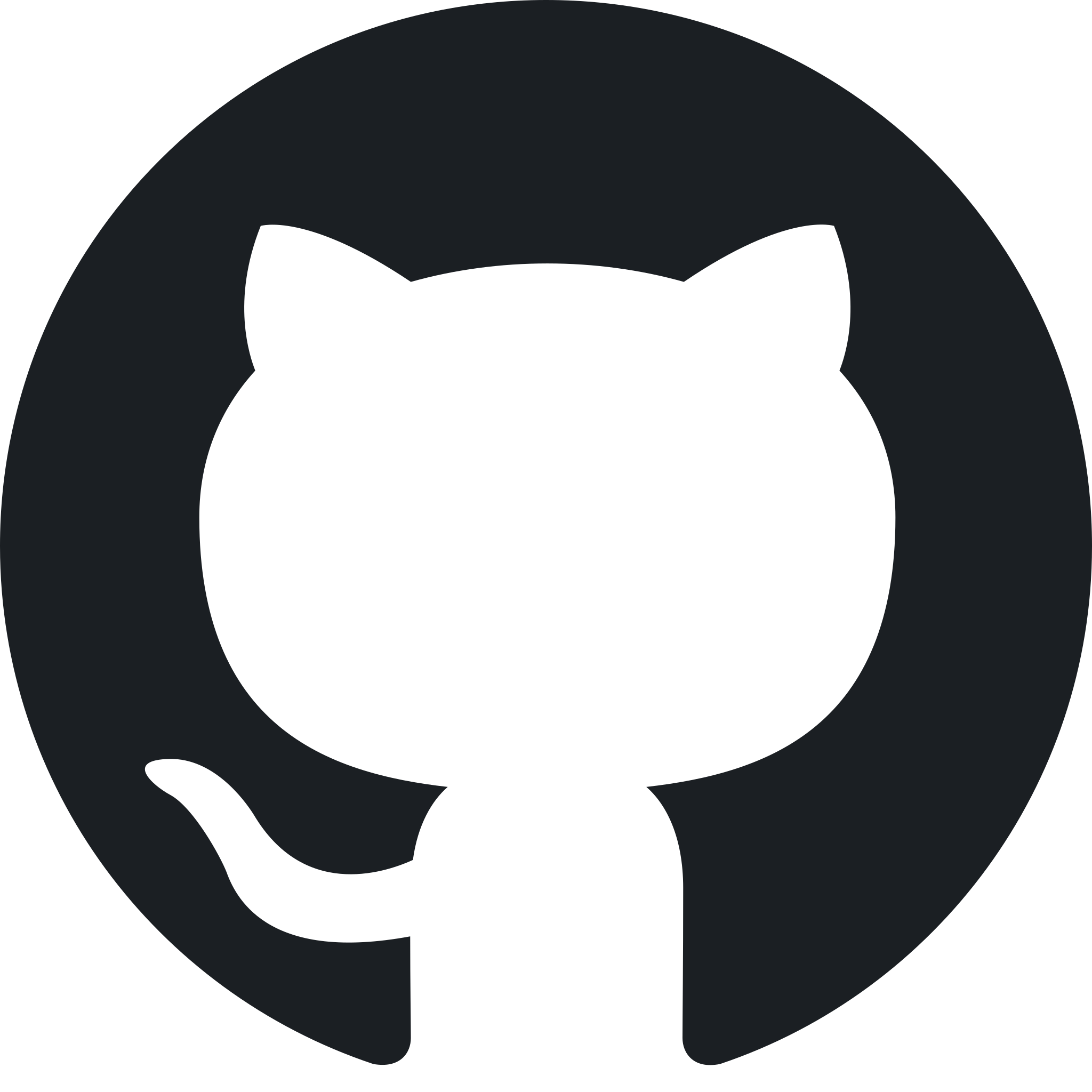} \small Code: \url{https://github.com/SalesforceAIResearch/MAS-Zero} 
}

\begin{document}

\maketitle

\begin{abstract}
Multi-agent systems (MAS) leveraging the impressive capabilities of Large Language Models (LLMs) hold significant potential for tackling complex tasks. However, most current MAS depend on manually designed agent roles and communication protocols. These manual designs often fail to align with the underlying LLMs' strengths and struggle to adapt to novel tasks.
Recent automatic MAS approaches attempt to mitigate these limitations but typically necessitate a validation set for tuning and yield static MAS designs lacking adaptability during inference, while also removing the flexibility to reduce to simpler systems. We introduce \ourframework, the first self-evolved, inference-time framework for automatic MAS design. \ourframework employs meta-level design to iteratively design, critique, and refine MAS configurations tailored to \emph{each problem instance}, without requiring a validation set. Critically, it enables dynamic problem decomposition and agent composition through meta-feedback on solvability and completeness, and reduction to simpler systems when appropriate. Experiments across reasoning (math and graduate-level QA), coding, and agentic (search-based) benchmarks, using both closed-source and open-source LLM backbones of varying sizes, demonstrate that \ourframework outperforms strong manual and automatic MAS baselines. It achieves substantial average accuracy improvements of up to 16.69\% on reasoning, 16.66\% on coding, and 5.45\% on agentic tasks, while maintaining cost efficiency.
\end{abstract}

\section{Introduction}


While standalone large language models (LLMs) have demonstrated strong performance across numerous tasks~\citep{deepseekr1,ke2025demystifying,vu2024foundational}, many problems remain too intricate for a single model to solve effectively~\citep{wang2024rethinkingboundsllmreasoning,guo2024large}. To tackle these challenges, the exploration of multi-agent systems (MAS) composed of multiple LLM agents has gained increasing traction among researchers~\citep{ke2025surveyfrontiersllmreasoning}.\footnote{Agents in a MAS can interact with external environmental tools e.g., search tools~\citep{jiang2024rag}, or collaborate with other agents to address tasks requiring diverse capabilities or multiple steps~\citep{liang2023encouraging,chen-etal-2024-reconcile}. This work focuses on the latter scenario, where each agent within the MAS is an LLM communicating with other LLM agents.} These agents often assume distinct \textit{roles}, such as generator or verifier \citep{reflexion_shinn2024}, engage in debates offering varied perspectives \citep{qian2025scalinglargelanguagemodelbased,wang2024mixtureofagentsenhanceslargelanguage}, and {\color{black}perform assigned sub-tasks \citep{li2025agentorientedplanningmultiagentsystems}}. 

A fundamental challenge in MAS lies in designing an effective connection and configuration of these agents to solve a given problem. {\color{black}Initially, MAS were handcrafted, with humans designing both agent roles and inter-agent communication protocols. However, MAS composed entirely of such manually designed configurations have faced issues such as poor problem specification and inter-agent misalignment~\citep{cemri2025multiagentllmsystemsfail,qiao2025benchmarkingagenticworkflowgeneration}, especially when the MAS agents are not specifically trained with such configurations.}

\begin{figure*}[t]
\centering
\includegraphics[width=0.9\textwidth]{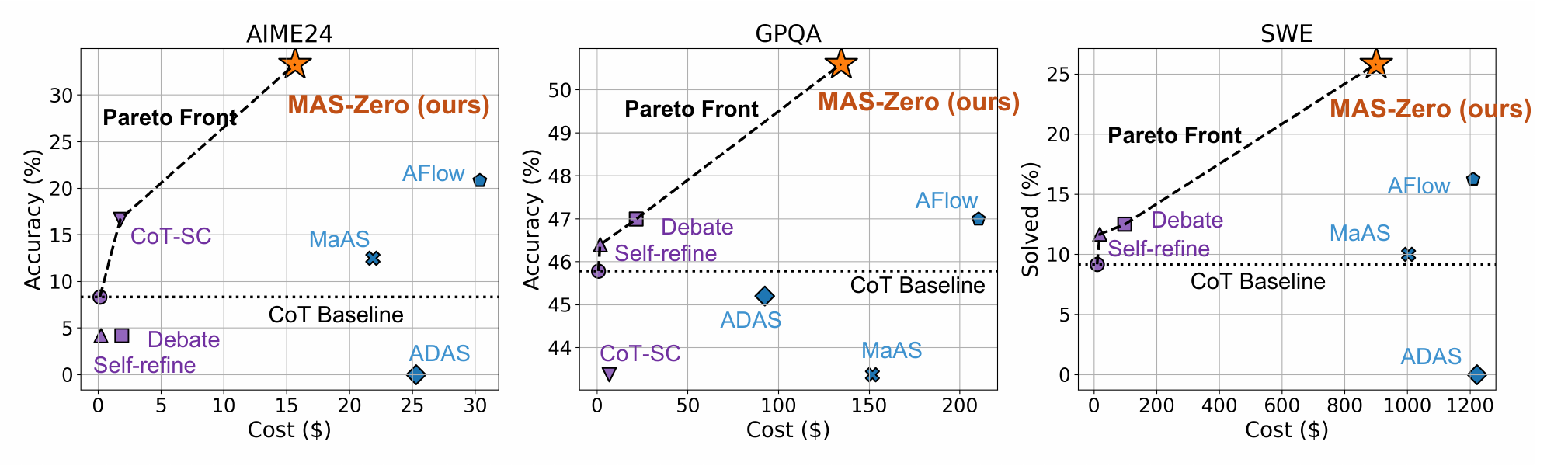}
\caption{{\small \color{black}
Scatter plots comparing the Pareto fronts of various {GPT-4o-based} systems on three benchmarks, including
\MyPurple{manual MAS}, \MyBlue{automatic MAS} and \MyOrange{\ourframework}. \ourframework delivers high performance at lower cost than comparable automatic MAS methods, establishing a new frontier for accuracy vs. cost trade-off. 
}
}
\label{fig.pareto_front}
\vspace{-1\baselineskip}
\end{figure*}

These shortcomings are understandable, as manually specifying a MAS is difficult when the human designer and the underlying LLMs are not well aligned. Moreover, manual approaches do not scale well to novel problems, especially as the problems become more complex. Recent work has explored automatic MAS design, 
but they have significant limitations: (1) Most rely on a ``training'' phase with labeled validation sets to tune configurations, which are often unavailable in real-world scenarios and may not generalize. This training, based solely on \textit{outcome} correctness, provides limited insight into the system's internal dynamics. 
(2) This reliance on validation sets often yields a \textit{fixed} architecture (i.e., one for the entire problem set) which lacks per-problem adaptability at test time. 
(3) Even worse, these methods eliminate critical dynamics: they cannot reduce to a simpler MAS or a single-agent system when such strategies would be stronger~\citep{huang2024large}, nor can they flexibly decompose a problem into smaller, more manageable sub-tasks.
This limitation is less apparent on simple tasks such as GSM8K and HumanEval, which are commonly used for MAS evaluation~\citep{hu2025automated,zhang2025multiagentarchitecturesearchagentic,zhang2024aflowautomatingagenticworkflow} and where most methods already perform well. On more challenging tasks (e.g., AIME24), however, the inability to revert or decompose becomes critical: as shown in Fig.~\ref{fig.pareto_front}, many baselines show little to no improvement over simple CoT, meaning the integrated system does not even outperform a single component.

To overcome these limitations, we argue that an effective automatic MAS should satisfy three core desiderata: \textbf{(1)} be dynamic enough to both decompose complex problems into smaller sub-tasks and reduce to a single agent or a simple MAS when a sophisticated MAS is \textit{not needed}; \textbf{(2)} learn the capabilities of the underlying LLMs, and automatically design MAS structures that are aligned with those capabilities;
and \textbf{(3)} support adaptivity at inference time, so that MAS designs can be tailored \textit{per problem instance} without relying on training or validation sets. To our knowledge, no existing automatic MAS framework satisfies all three desiderata simultaneously. In this work, we propose a novel automatic inference-time MAS optimization framework, called \textit{\textbf{MAS-ZERO}}, which designs MAS with \emph{zero} supervision, while satisfying all the aforementioned desiderata. In particular, \ourframework introduces a \textbf{meta-agent} that iteratively learns the capabilities of individual agents and their combinations, and refines the MAS design accordingly, thus operating at the MAS-level rather than the agent level (hence “meta”). The meta-agent also verifies candidate answers drawn from both refined MAS designs and simpler MAS or single-agent systems, ensuring the dynamic reduction capability. This process operates entirely at test time, allowing for unique MAS designs \textit{per-problem}.
 
\begin{figure*}[t]
\centering
\includegraphics[width=0.8\textwidth]{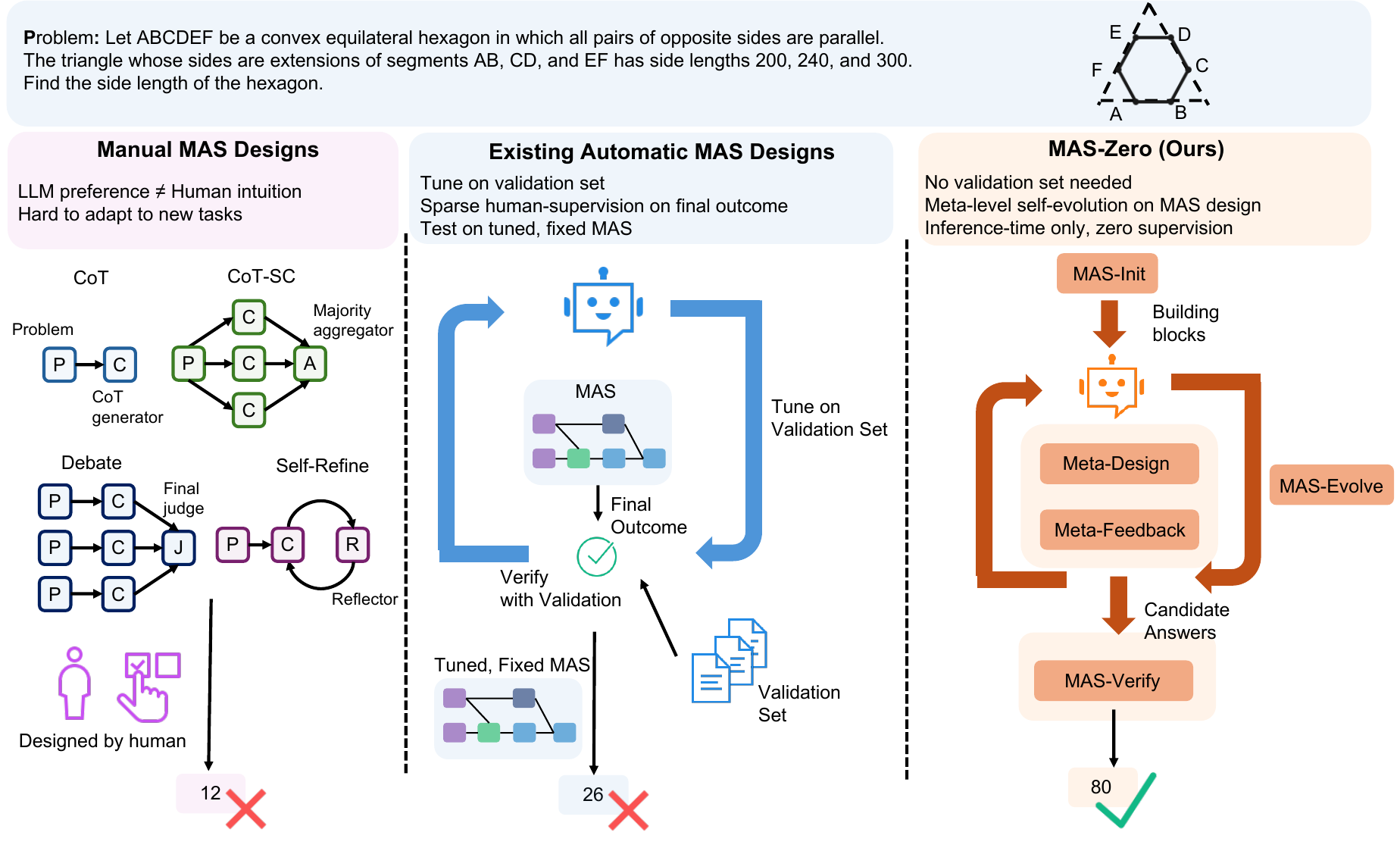}
\vspace{-1\baselineskip}
\caption{\small Conceptual comparison of \ourframework, with existing automatic and manual MAS designs. \ourframework avoids tuning MAS on validation set by maintaining a self-evolving process that iteratively designs and evaluates task-specific MAS at inference time. 
}
\label{fig.conceptual_contrast}
\vspace{-1\baselineskip}
\end{figure*}

To achieve this, \ourframework tasks the meta-agent to iteratively \textit{design} and \textit{critique} the MAS, maintain an \textit{experience library}, \textit{refine} the design based on accumulated experience, and ultimately \textit{verify} the candidate answers. Fig.~\ref{fig.conceptual_contrast} illustrates a conceptual overview and contrasts \ourframework with both automatic and manual MAS designs. Specifically, \ourframework involves three key steps:


\begin{itemize}[leftmargin=*,noitemsep,topsep=2pt]
    \item \textbf{Initializing building blocks (MAS-Init)}: \ourframework starts with established single-agent (e.g., CoT, Self-Consistency) and simple human-designed MAS strategies (e.g., Debate, Self-Refine), executing each to generate initial outputs that seed later steps.
    
    \item \textbf{Self-evolving with iterative refinement (MAS-Evolve)}: The meta-agent iteratively \textit{designs} and \textit{critiques} MAS configurations, guided by feedback on \textit{solvability} and \textit{completeness}, while accumulating prior designs and feedback in an experience library for continual refinement.
    
    \item \textbf{Selecting the best candidate with self-verification (MAS-Verify)}: From the pool of outputs, including both building blocks and refined MAS iterations, the meta-agent verifies and selects the most reliable solution, dynamically choosing between complex MAS and simpler strategies.
\end{itemize}

Evaluations across three challenging domains—reasoning (math and graduate-level QA), coding, and agentic (search-based), using both closed-source and open-source LLM backbones of varying sizes (including GPT-4o, 32B, and 70B models) demonstrate that \ourframework consistently outperforms strong manual and automatic MAS baselines. It achieves substantial average accuracy improvements of up to 16.69\% on reasoning, 16.66\% on coding, and 5.45\% on agentic tasks. It also consistently lies on the Pareto frontier of accuracy and cost (Fig.~\ref{fig.pareto_front}). While the inference-time mechanism incurs higher token usage during testing, it avoids expensive validation-time optimization and shifts the design effort to the testing phase, where it can flexibly handle new tasks, and often be more effective \citep{agrawal2025gepareflectivepromptevolution}. Such a trade-off has demonstrated significantly improved answers in this work and strong potential in the literature \citep{liu2024groupdebate}. We believe that \ourframework provides a complementary alternative for the MAS community, especially in scenarios where adaptability and generality outweigh the need for minimal token usage.
In summary, our key contributions are:

\begin{itemize}[leftmargin=*,noitemsep,topsep=2pt]
    \item We introduce \ourframework, to our knowledge, the \textbf{first inference-time-only} automatic MAS design framework. It works in a fully self-evolved way by learning from the behavior of the underlying LLM agents \textit{at inference-time}, enabling per-instance adaptivity with \textit{zero} supervision.
    \item We present \textbf{a new SoTA automatic MAS system} that achieves substantial performance gains over both manually designed and strong automatic baselines, while remaining cost-efficient and Pareto-optimal across diverse LLMs and domains.\footnote{We will open-source the data, code, and leaderboard for all components upon acceptance.}
    \item We conduct a comprehensive evaluation of \ourframework across diverse domains and LLMs, presenting \textbf{key insights}. For example, single-agent or simple MAS configurations can be surprisingly strong, in some cases even outperforming more sophisticated MAS designs. Crucially, \ourframework is the only system that can \textit{dynamically revert} to these simpler yet effective strategies, ensuring that such strengths are not overlooked.
\end{itemize}

\section{Related Work}
\label{sec.related_work}





\vspace{-0.2em}\noindent\textbf{Manual MAS design.} 
Building on the success of single-agent systems {\color{black}(e.g., CoT~\citep{wei2022chain}, self-consistency (CoT-SC)~\citep{wang2023selfconsistency})}, studies have shown that grouping multiple LLM agents into a MAS can substantially improve individual agent performance. To this end, a variety of manual-designed MAS approaches have been proposed~\citep{xu2025expertpromptinginstructinglargelanguage,zheng2024stepbackevokingreasoning,lu2025intelligentgoexplorestandingshoulders}, including LLM debate~\citep{du2023improvingfactualityreasoning}, and self-refine~\citep{madaan2024self}.
However, as discussed previously, these manual designs often suffer from limited adaptability and scalability, and their rigid structures may fail to align with the underlying strengths of LLMs.




\noindent\textbf{Automatic MAS design.} 
Recent work on automatic MAS design typically require validation set. We broadly categorize them into two families: \textbf{(1) val-pruning} starts with a fully connected, pre-defined graph of LLM agents or human-designed blocks and prune it based on validation performance.
For example, MASS~\citep{zhou2025multiagentdesignoptimizingagents} uses rejection sampling, and MaAS~\citep{zhang2025multiagentarchitecturesearchagentic} extends MASS with a question-wise masking mechanism to adapt subnetworks. However, their adaptability remains limited as the core MAS structure is constrained by the pre-defined structure, which is suboptimal for many tasks; \textbf{(2) val-generation} leverages a meta-agent LLM to generate MAS from scratch, offering greater flexibility in defining novel agents and architectures compared to pre-defined structures. However, this expanded design space presents significant learning challenges. Recent efforts including ADAS~\citep{hu2025automated} and AFlow~\citep{zhang2024aflowautomatingagenticworkflow} frame MAS generation as a code generation task. ADAS stores and searches historical designs based on validation performance, while AFlow enhances this with Monte Carlo Tree Search. Our framework also represents MAS as executable code but differs fundamentally in several ways: instead of relying on potentially unreliable validation sets, \ourmodel uses a self-evolving process at inference time to learn the capabilities of agents for meta-level design.
It further integrates question decomposition into MAS design, enabling MAS to be constructed and refined at the sub-task level. Finally, \ourframework can dynamically revert to simpler building blocks when they are sufficient. These capabilities are not supported by existing automatic MAS systems. More methods like DyLAN~\citep{liu2024dynamicllmpowered} are discussed in App. \ref{ap.related_work}.

\section{\ourframework Framework}
\label{sec.framework}
\vspace{-0.5em}



\begin{figure*}[t]
\centering
\includegraphics[width=0.8\textwidth]{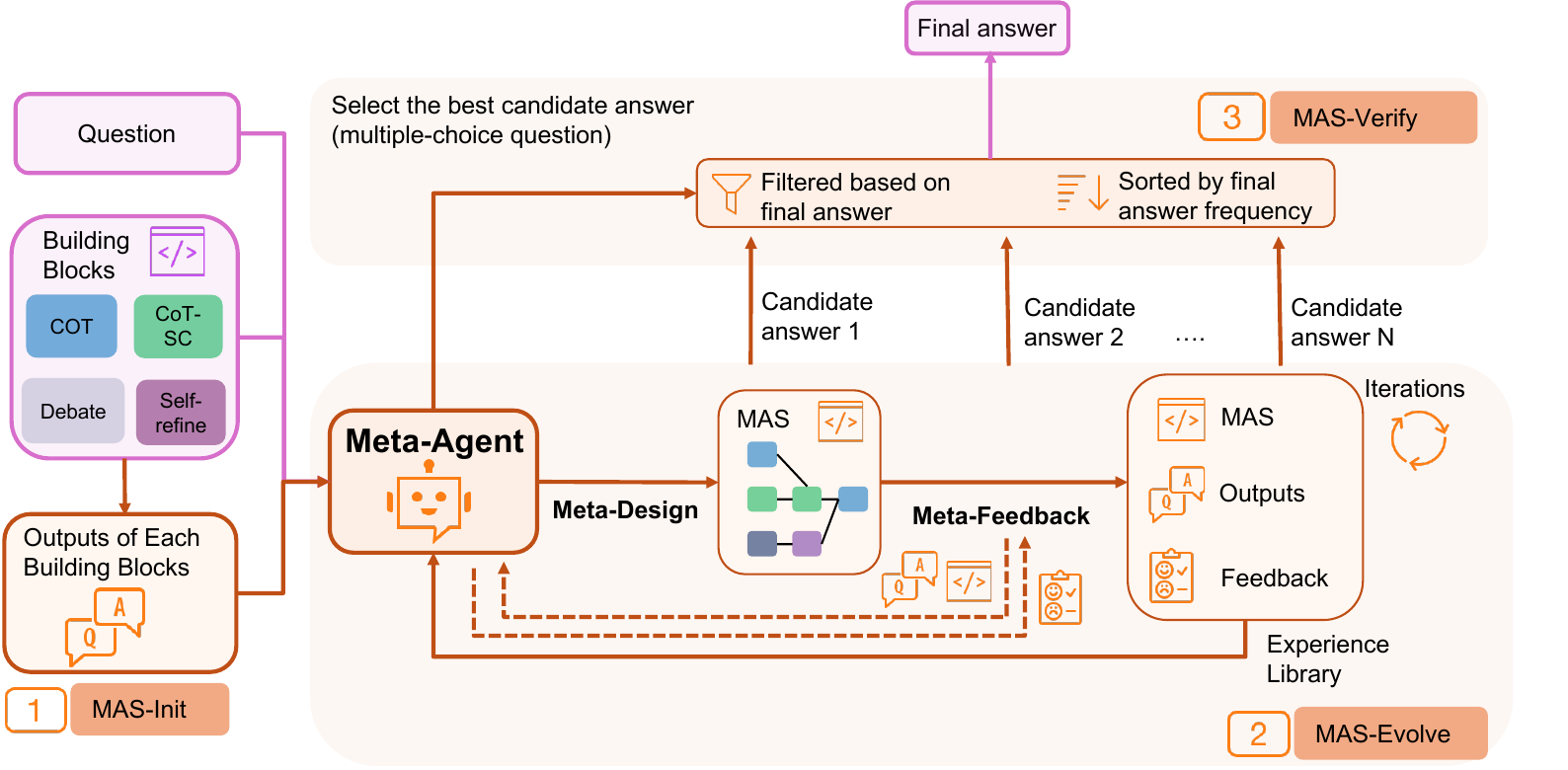}
\vspace{-0.8\baselineskip}
\caption{\small \ourframework overview. 
\MyPurple{Purple} highlights the given input and final output. \MyOrange{Orange} highlights the components and steps in \ourframework. \underline{Dashed arrows} indicate the information flow \textit{within} Meta-feedback. \ourframework takes as inputs the question and building blocks, and solves the task in three key steps: \textbf{MAS-Init} (Sec.~\ref{sec.mas_blocks}), \textbf{MAS-Evolve} (Sec.~\ref{sec.meta_iter}) and \textbf{MAS-Verify} (Sec.~\ref{sec.self_verify}). 
}
\label{fig.framework_overview}
\vspace{-1\baselineskip}
\end{figure*}

As shown in Fig.~\ref{fig.framework_overview}, \ourframework first conducts \textbf{MAS-Init} (Sec.~\ref{sec.mas_blocks}), where it executes each of the given building blocks. It then takes a question, the building blocks, and the outputs of each building block as inputs, ultimately producing the final answer. These inputs are processed by the central \textbf{meta-agent}, which orchestrates both the \textbf{MAS-Evolve} (Sec.~\ref{sec.meta_iter}) and \textbf{MAS-Verify} (Sec.~\ref{sec.self_verify}) steps. Importantly, the whole process functions \textit{without} prior knowledge or internal details of the underlying LLM agents. All steps are implemented through prompting and require only black-box access to LLM generation, making \ourframework broadly applicable to any LLM without requiring fine-tuning or internal modifications. The corresponding pseudocode is provided in App.~\ref{ap.algo}.


\subsection{MAS-Init}
\label{sec.mas_blocks}
\vspace{-0.5em}

MAS-Init serves as the entry point of \ourframework by executing a set of predefined \textit{building blocks}. These blocks correspond to established human-designed strategies (CoT, CoT-SC, Debate, and Self-Refine in this work) implemented as executable code. Given a question, MAS-Init runs each block to generate initial candidate solutions. These blocks and their outputs are used as: (1) input to the meta-agent for grounding the MAS design (Sec.~\ref{sec.meta_iter}), and (2) candidate answers that can be selected by the MAS-Verify (Sec.~\ref{sec.self_verify}), enabling dynamic reduction to  simpler MAS or single-agent systems.


\subsection{MAS-Evolve}
\label{sec.meta_iter}
\vspace{-0.5em}

Given the question, the building blocks, and their outputs from MAS-Init, the meta-agent begins to design the MAS. Initially, it has no knowledge of the underlying LLM agents' internal capabilities and may produce suboptimal designs. We propose an iterative process in which the meta-agent gradually learns the strengths of the component agents and refines its designs.
This process alternates between two phases: \textbf{(1) meta-design} (Sec.~\ref{sec.meta_design}), where the meta-agent decomposes the  question into sub-tasks and proposes a MAS based on the building blocks and any accumulated \textit{experience} from prior iterations; \textbf{(2) meta-feedback} (Sec.~\ref{sec.meta_feedback}), where the meta-agent reviews the proposed MAS and sub-tasks using intermediate outputs to assess their solvability and completeness, and then generates targeted feedback. The MAS, its intermediate outputs and feedback, are stored in an \textit{experience library} that informs subsequent iterations. Through this cycle, the meta-agent progressively adjusts decompositions and configurations, yielding continual improvement without external supervision.








\subsubsection{Meta-Design}
\label{sec.meta_design}
\vspace{-0.5em}
The goal of this phase is to design a candidate MAS for the given task, which will then be reviewed in the next phase. Unlike existing work that tackles complex problems all at once, \ourframework explicitly decomposes the original question into \textit{manageable} yet \textit{interdependent} sub-tasks. This decomposition not only breaks down complex problems into smaller parts but also creates opportunities to assign sub-task level MAS (i.e., \textit{sub-MAS}) tailored to different components of the problem.
For each sub-task, the meta-agent assigns a sub-MAS by modifying connections between given building blocks or adjusting their parameters (e.g., temperature, number of debate rounds, etc.). This deliberate design, informed by our preliminary experiments, balances exploration with improvement: the meta-agent is free to analyze questions, sub-tasks and assigned sub-MAS, but it should not arbitrarily invent new agents or blocks, nor prune the architecture without grounding in the provided building blocks.\footnote{To support code generation, we provide a template with utility functions and apply sanity checks (syntax validation and field consistency; see App.~\ref{ap.code_template}).}

\subsubsection{Meta-Feedback} 
\label{sec.meta_feedback}
\vspace{-0.5em}

\noindent\textbf{MAS and intermediate outputs.} Given the design produced in the meta-design phase, the meta-feedback phase reviews the MAS and generates feedback. Since a MAS is executable code, it can be run to obtain outputs, but relying only on the final answer is often sparse and uninformative. \ourframework instead exploits the intermediate outputs, incorporating both \textit{sub-task level} outputs from sub-MAS and \textit{agent-level} outputs from individual LLMs. By jointly analyzing the final answer and these fine-grained signals, the meta-agent gains a much richer view of strengths and weaknesses across the MAS. Concretely, with the code-based representation, each sub-MAS is executed to solve its sub-task, producing intermediate outputs at two levels: the \textit{sub-task (sub-MAS)} level and the \textit{agent} level. These outputs form the basis for evaluation against the key criteria introduced below.


\noindent\textbf{Criteria.}
Given the above sub-task and agent level outputs, \ourframework evaluates \textit{solvability} and \textit{completeness}. The meta-agent is given agency in determining each metric:

\begin{itemize}[leftmargin=*,noitemsep,topsep=2pt]
    \item \textbf{Solvability} requires that each sub-task be \textit{independently} and \textit{completely} solvable by its sub-MAS, ensuring that every sub-task yields reliable outputs.\footnote{To further aid the meta-agent, we allow each agent to output a special token, \texttt{[TOO HARD]}, if it determines that the assigned sub-task is beyond its current capabilities.} 
    
    \item \textbf{Completeness} requires that the complete set of sub-tasks covers all necessary information from the original input, ensuring that their answers can produce a correct and comprehensive aggregated answer to the original task. While an individual sub-task may address only part of the necessary content, all critical information must be processed and used at some point in the MAS.
\end{itemize}    

\noindent\textbf{Generating feedback.}
Based on the solvability and completeness, the meta-agent generates targeted natural language feedback on specific aspects of the MAS that may require revision. For example, if a sub-task is identified as not solvable, the feedback should suggest either further decomposing it or updating the corresponding sub-MAS in the next iteration. Conversely, if a sub-task is considered solvable, the feedback should indicate that it and its sub-MAS remain unchanged. Similarly, if the union of sub-tasks is found to miss necessary information, the feedback should recommend refining the decomposition of the original problem to incorporate the missing elements. Overall, this feedback guides subsequent meta-design iterations, allowing the overall system to iteratively converge toward an effective decomposition and MAS.

\subsubsection{Storing the Experience and Refining the design}
After the first meta-design (Sec.~\ref{sec.meta_design}), meta-feedback is collected (Sec.~\ref{sec.meta_feedback}). The MAS, its intermediate outputs, and the associated feedback are stored as \textit{experience} in an \textit{experience library}. In each subsequent iteration, meta-design is performed again, now with experience from the library provided as additional context to drive self-evolution. Through this process, the meta-agent dynamically adapts its decomposition strategy and sub-MAS assignments across iterations. This iterative accumulation of experience gives \ourframework a persistent memory, enabling it to leverage knowledge from past iterations and build a stronger foundation for continual improvement. 

As in many other self-evolving frameworks \citep{gao2025surveyselfevolvingagentspath}, the meta-design and meta-feedback signals may be imperfect and ultimately depend on the underlying LLM. Nevertheless, we find empirically that \ourframework allows initially imperfect designs to be progressively improved (Fig.~\ref{fig.mas_moment}), and that our curated instruction design produces strong feedback—outperforming a simple ensemble alternative (Sec.~\ref{sec.ablation}). We view these results as a promising starting point and hope they inspire further research in advancing strategies for iterative MAS refinement

\subsection{MAS-Verify}
\label{sec.self_verify}
\vspace{-0.5em}


\noindent\textbf{Collecting candidate answers.} 
At each iteration of MAS-Evolve, the MAS is executed to produce intermediate outputs and a \textit{candidate answer} (including both the chain-of-thought and the final answer). After multiple rounds, \ourframework must determine which candidate answer is the most reliable and complete. Importantly, the pool of candidate answers includes \textit{not only} those generated in each iteration of MAS-Evolve \textit{but also} the outputs of the basic building blocks from MAS-Init.
This design allows the meta-agent to select between them, leveraging the strong performance of simple strategies when they suffice, while also exploiting the complex MAS when needed.



\noindent\textbf{Verifying answers.} Relying on the last iteration (or any single iteration) is suboptimal due to stochastic LLM outputs and ongoing MAS refinement (ablations in Sec.~\ref{sec.ablation}). Instead, \ourframework formulates verification as a \textit{selection} problem and tasks the meta-agent with selecting the most coherent and correct output from the set of candidate answers, which is often more tractable than independently scoring each output~\citep{gu2025surveyllmasajudge,zhou2025evaluating}, especially for challenging questions where correctness is hard to assess in isolation. 
Specifically, \ourframework first \textit{ranks} candidates by their final answer frequency. This acts as a prior favoring majority responses, a strategy shown to be effective in prior work~\citep{wang2023selfconsistency}. It then \textit{filters} out clearly invalid answers (e.g., not among the given options). Finally, it \textit{selects} the best answer from the remaining candidates.

\section{Experiments}
\label{sec:experiments}


\vspace{-0.5em}
\noindent\textbf{Setup.} 
We consider both the closed-source \textbf{GPT-4o}~\citep{gpt4o} (web-search version for agentic tasks) and the open-source LLMs,
\textbf{Llama3.3-70B-inst}~\citep{grattafiori2024llama3herdmodels} and \textbf{Qwen2.5-32B-inst}~\citep{qwen2025qwen25technicalreport}. To fairly evaluate how well \ourmodel performs relative to the underlying LLM used to construct the MAS, we always use the \textit{same} LLM for both the meta-agent and individual agents (heterogeneous settings in Sec.~\ref{sec.ablation}).  We use the same prompt template for all the tasks (App.~\ref{ap.prompt_detail} and \ref{ap.code_template}) and conduct 5 MAS-Evolve iterations (the maximum permitted by context-length). Together with the 4 building blocks in MAS-Init, this yields 9 candidate answers, from which the meta-agent selects one final answer with MAS-Verify.

\noindent\textbf{Benchmarks.} 
We consider 2 \textbf{reasoning} benchmarks across math and science: \textbf{AIME24}~\citep{aime2024} and GPQA-diamond \textbf{(GPQA)}~\citep{rein2023gpqagraduatelevelgoogleproofqa} (graduate-level QA), 1 \textbf{coding} benchmark SWE-Bench-Lite-Oracle \textbf{(SWE)}~\citep{jimenez2024swebench},\footnote{Note that MAS-Verify does not apply to SWE, as correctness in SWE is determined directly by the compiler.} and 2 search-based \textbf{agentic} benchmarks: \textbf{BrowseComp} \citep{wei2025browsecompsimplechallengingbenchmark} and \textbf{Frames} \citep{krishna2025factfetchreasonunified}.\footnote{Benchmark statistics and more implementation details can be found in App.~\ref{ap.benchmark_baseline}.} Existing automatic MAS methods largely restrict their evaluations to relatively simple reasoning tasks. To our knowledge, \ourframework is the first to conduct evaluations on challenging reasoning, coding and agentic tasks. 

\noindent\textbf{Baselines.} 
We include 2 widely used \textbf{single-agent} baselines: \textbf{CoT}~\citep{wei2022chain} and self-consistency \textbf{(CoT-SC)}~\citep{,wang2023selfconsistency}; 6 \textbf{manual MAS} baselines: \textbf{Debate}~\citep{du2023improvingfactualityreasoning},  \textbf{Self-refine}~\citep{madaan2024self}, \textbf{ReConcile} \citep{chen-etal-2024-reconcile}, \textbf{MAD} \citep{liang2023encouraging}, \textbf{SPP} \citep{wang2023unleashing} and \textbf{DyLAN} \citep{liu2024dynamicllmpowered}. Note that CoT, CoT-SC, Debate and Self-Refine also serve as the building blocks in MAS-Init, allowing us to clearly observe how our system improves upon the initial configurations. For \textbf{automatic MAS}, we include 3 strong methods: \textbf{val-pruning} \textbf{MaAS}~\citep{zhang2025multiagentarchitecturesearchagentic} and \textbf{val-generation} \textbf{ADAS}~\citep{hu2025automated}  and \textbf{AFlow}~\citep{zhang2024aflowautomatingagenticworkflow}. We also include the latest \textbf{training-based} method \textbf{MAS-GPT} \citep{ye2025masgpttrainingllmsbuild}.



\vspace{-0.5em}
\subsection{Overall Results}
\label{sec.overall_results}
\vspace{-0.5em}

\begin{table*}[t]
\setlength{\fboxsep}{1pt}
\centering
\setlength{\tabcolsep}{2pt}
\resizebox{0.8\textwidth}{!}{
\begin{tabular}{l>{\color{gray}}l>{\color{gray}}ll>{\color{gray}}l>{\color{gray}}ll>{\color{gray}}l>{\color{gray}}ll}
\toprule
\textbf{LLMs} & \multicolumn{3}{c}{\textbf{GPT-4o}} & \multicolumn{3}{c}{\textbf{Llama3.3-70B}} & \multicolumn{3}{c}{\textbf{Qwen2.5-32B}} \\
\textbf{Methods} & \textbf{AIME24} & \textbf{GPQA} & \multicolumn{1}{c}{\textbf{Avg.}} & \textbf{AIME24} & \textbf{GPQA} & \multicolumn{1}{c}{\textbf{Avg.}} & \textbf{AIME24} & \textbf{GPQA} & \multicolumn{1}{c}{\textbf{Avg.}} \\
\toprule
\cellcolor{cyan!20} CoT & 8.33 & 45.78 & 27.06\upgreen{14.91} & 16.67 & 50.60 & 33.63\upgreen{11.32} & 12.50 & 50.00 & 31.25\upgreen{9.24} \\
\cellcolor{cyan!20} CoT-SC & 16.67 & 43.37 & 30.02\upgreen{11.95} & 29.17 & 51.20 & 40.18\upgreen{4.77} & 16.67 & 49.40 & 33.04\upgreen{7.46} \\
\cellcolor{purple!20} Debate & 4.17 & 46.99 & 25.58\upgreen{16.39} & 20.83 & 50.60 & 35.72\upgreen{9.24} & 8.33 & 49.40 & 28.87\upgreen{11.63} \\
\cellcolor{purple!20} Self-Refine & 4.17 & 46.39 & 25.28\upgreen{16.69} & 29.17 & 54.22 & 41.69\upgreen{3.26} & 16.67 & 50.60 & 33.64\upgreen{6.86} \\
\cellcolor{purple!20} ReConcile & 12.50 & 48.43 & 30.47\upgreen{11.50} & 33.33 & 47.17 & 40.25\upgreen{4.71} & 12.50 & 47.17 & 29.84\upgreen{10.66} \\
\cellcolor{purple!20} MAD & 13.89 & 52.01 & 32.95\upgreen{9.02} & 29.17 & 52.61 & 40.89\upgreen{4.07} & 16.67 & 43.57 & 30.12\upgreen{10.37} \\
\cellcolor{purple!20} SPP & 9.72 & 49.80 & 29.76\upgreen{12.21} & 26.39 & 46.18 & 36.29\upgreen{8.67} & 19.44 & 42.77 & 31.11\upgreen{9.39} \\
\cellcolor{purple!20} DyLAN & 11.11 & 46.99 & 29.05\upgreen{12.92} & 29.17 & 41.57 & 35.37\upgreen{9.59} & 20.83 & 42.57 & 31.70\upgreen{8.79} \\
\midrule
\cellcolor{blue!10} MaAS & 12.50 & 43.37 & 27.94\upgreen{14.03} & 33.33 & 43.98 & 38.65\upgreen{6.30} & 20.83 & 46.99 & 33.91\upgreen{6.58} \\
\cellcolor{blue!30} ADAS & $\times$ & 45.20 & $\times$ & 8.30 & 53.60 & 30.95\upgreen{14.01} & 12.50 & 47.00 & 29.75\upgreen{10.74} \\
\cellcolor{blue!30} AFlow & 20.83 & 46.99 & 33.91\upgreen{8.05} & 33.33 & 47.59 & 40.46\upgreen{4.49} & 33.33 & 46.39 & 39.86\upgreen{0.63} \\
 \midrule
\cellcolor{green!20} MAS-GPT & 13.89 & 43.98 & 28.94\upgreen{13.03} & 26.39 & 40.00 & 33.20\upgreen{11.76} & 23.61 & 37.35 & 30.48\upgreen{10.01} \\
 \midrule
\cellcolor{orange!20} \textbf{\ourframework} & 33.33 & 50.60 & \textbf{41.97} & 37.50 & 52.41 & \textbf{44.96} & 29.17 & 51.81 & \textbf{40.49} \\
\bottomrule
\end{tabular}
}
\caption{\small Reasoning tasks results. ``$\times$'' indicates 0\% accuracy for MAS selected using the validation set. ``${\color{darkgreen}\uparrow}$'' denotes the difference (improvement) that \ourframework achieves compared to the baselines. Highlighting indicates \SingleAgent{single-agent}, \ManualMAS{manual MAS},\ValidationPrune{val-pruning automatic MAS }, \ValidationGen{val-generation automatic MAS}, \TrainingMAS{training-based automatic MAS} and \OurMethod{our method}. To fairly compare with validation-based baselines, we split each benchmark’s original test set into 20\% for validation and 80\% for testing. For methods do not use validation sets (including \ourmodel), we evaluate on the same 80\% split. Standard deviations are given in App. \ref{ap.error_bar}.
}
\label{tab.reasoning_results}
\vspace{-1\baselineskip}
\end{table*}

Tables~\ref{tab.reasoning_results}-\ref{tab.agentic_results} show the results for reasoning, coding and agentic tasks across applicable LLMs and benchmarks. On average, {\ourmodel achieves the \textit{best} performance across all LLMs and domains}. Below, we summarize the additional takeaways from the comparison:


\noindent\textbf{{Reasoning Tasks}.} From Table~\ref{tab.reasoning_results}, we observe that (1) \ourmodel \textit{consistently outperforms} all \textit{automatic MAS} methods. Across all LLM backbones and benchmarks, it surpasses SoTA baselines, exceeding the strongest baseline, AFlow, by 13.03\% on average with GPT-4o as the backbone. The only instance where it falls behind is on AIME24 with the Qwen backbone, where it underperforms AFlow by merely one sample (out of 24 total). Notably, ADAS fails completely on AIME24 (0\% accuracy), despite having access to a validation set, underscoring the unreliability of validation-based baselines. (2) \ourmodel also \textit{consistently outperforms} strong single-agent and manual MAS, with only two exceptions: GPQA with MAD using GPT-4o, and GPQA with Self-Refine using Llama. 

\begin{wraptable}{r}{0.35\textwidth}
\vspace{-1\baselineskip}
\setlength{\fboxsep}{1pt}
\centering
\setlength{\tabcolsep}{2pt}
\resizebox{0.3\textwidth}{!}{
\begin{tabular}{lll}
\toprule
\textbf{LLMs} & \textbf{GPT-4o} & \textbf{Llama3.3} \\
\textbf{Methods} & \textbf{SWE} & \textbf{SWE} \\
\toprule
\cellcolor{cyan!20} CoT & 9.17\upgreen{16.66} & 2.92\upgreen{13.82}  \\
\cellcolor{purple!20} Debate & 12.50\upgreen{13.33} & 6.67\upgreen{10.07} \\
\cellcolor{purple!20} Self-Refine & 11.67\upgreen{14.16} & 1.67\upgreen{15.07} \\
\cellcolor{blue!10} MaAS & 10.00\upgreen{15.83} & 5.00\upgreen{11.74}  \\
\cellcolor{blue!30} AFlow & 16.25\upgreen{9.58} & 6.67\upgreen{10.07} \\
\midrule
\cellcolor{orange!20} \textbf{\ourframework} & \textbf{25.83} & \textbf{16.74} \\
\bottomrule
\end{tabular}
}
\caption{\small SWE results. Methods that cannot be adapted to SWE are not included. 
Qwen is not included due to its small maximum context length (32K). 
}
\label{tab.coding_results}
\vspace{-1.2\baselineskip}
\end{wraptable}

Alarmingly, several automatic MAS baselines underperform manual MAS across multiple benchmarks. For example, CoT and CoT-SC consistently outperform MAS-GPT, ADAS, and MaAS. This further highlights the necessity of MAS-Init in \ourframework, as it allows the system to dynamically revert to simpler strategies when a sophisticated MAS is not needed.

\noindent\textbf{{Coding Tasks}.} Similar to reasoning tasks, in Table~\ref{tab.coding_results} we observe \ourframework clearly outperforms  single-agent, manual and automatic MAS. Notably, it comes with 58\% (GPT-4o) and 149\% (Llama) relative gains over the strongest baseline AFlow. These margins exceed those observed in reasoning tasks, highlighting the  effectiveness of \ourframework in challenging tasks.

\begin{wraptable}{r}{0.45\textwidth}
\vspace{-1\baselineskip}
\centering
\setlength{\tabcolsep}{2pt}
\resizebox{0.43\columnwidth}{!}{
\begin{tabular}{lccc}
    \toprule
    \textbf{LLM} & \multicolumn{3}{c}{\textbf{GPT-4o w/ search}} \\
    \textbf{Methods} & \textbf{BrowseComp} & \textbf{Frames} & \multicolumn{1}{c}{\textbf{Avg.}} \\
    \toprule
    CoT & 3.97 & 59.76 & 31.86\upgreen{5.45} \\
    CoT-SC & 8.66 & 63.58 & 36.12\upgreen{1.19} \\
    Debate & 3.94 & 70.45 & 37.19\upgreen{0.12} \\
    Self-Refine & 5.51 & 67.89 & 36.70\upgreen{0.61} \\
    \midrule
    \ourframework & 9.45 & 65.18 & \textbf{37.31} \\
    \bottomrule
\end{tabular}
}
\caption{Results on agentic tasks. 
}
\label{tab.agentic_results}
\vspace{-1\baselineskip}
\end{wraptable}

\textbf{Agentic Tasks.} We use GPT-4o with search as individual agent, which can query the internet and conduct multi-turn autonomous reasoning internally (meta-agent is still GPT-4o). From Table \ref{tab.agentic_results}, we see that on average, \ourframework continues to improve upon the basic building blocks.  On Frames, \ourframework underperforms Debate and Self-Refine. We speculate that when the search agent makes mistakes, those errors are grounded in retrieved content, making them more difficult to detect during MAS-Verify, leading to incorrect judgments. This highlights the importance of further strengthening the verifier in MAS-Verify (see Sec.~\ref{sec.ablation} for more analysis).


\noindent\textbf{Cost-efficiency.}
Fig.~\ref{fig.pareto_front} shows the trade-off between performance and cost for GPT-4o across three benchmarks. Cost is computed using the official OpenAI API pricing\footnote{More details are given in App.~\ref{ap.cost}} and includes both ``training'' (if any) and test-time usage. We observe that \textbf{\ourmodel lies on the Pareto front across all three datasets}. It is significantly more cost-efficient than AFlow, MaAS, and ADAS, with the lone exception of ADAS on GPQA, where the cost increase comes with a 12\% accuracy improvement. Of automatic MAS frameworks, \ourmodel delivers the highest performance at relatively low cost. While it is expected that automatic MAS methods incur higher costs than manual baselines, \ourmodel delivers substantially better performance, making the trade-off highly favorable.

\subsection{Further Analysis and Ablations}\label{sec.ablation}
\vspace{-0.5em}


While Sec.~\ref{sec.overall_results} establishes the overall effectiveness of \ourframework across domains and LLMs, here we analyze the role of the \textbf{meta-agent} and \textbf{each of the three steps} through a series of targeted ablations. The results, detailed below, show that a capable meta-agent consistently enhances performance and that all three steps contribute meaningfully and complementarily to the final improvements.






\begin{wraptable}{r}{0.4\textwidth}
\vspace{-1\baselineskip}
\centering
\setlength{\tabcolsep}{2pt}
\resizebox{0.4\columnwidth}{!}{
\begin{tabular}{lccc}
\toprule
\textbf{LLM} & \multicolumn{3}{c}{\textbf{o3-mini}} \\
\textbf{Methods} & \multicolumn{1}{c}{\textbf{AIME24}} & \multicolumn{1}{c}{\textbf{GPQA}} & \multicolumn{1}{c}{\textbf{Avg.}}\\
\midrule
CoT & 70.00 & 72.22 & 71.11\upgreen{12.27} \\
CoT-SC & 80.00 & 72.73 & 76.36\upgreen{7.02} \\
Debate & 86.67 & 77.78 & 82.22\upgreen{1.16} \\
Self-Refine & 76.67 & 74.24 & 75.45\upgreen{7.93} \\
\midrule
\ourframework & 90.00 & 76.77 & \textbf{83.38} \\
\bottomrule
\end{tabular}
}
\caption{\small \ourframework with stronger agents.}
\label{tab.stronger_agent}
\end{wraptable}

\noindent\textbf{Diverse meta-agents.} While \ourframework shows strong performance across various LLMs, we further examine whether weaker or stronger LLMs can effectively serve as meta-agents. For stronger LLM, we conduct experiments with a reasoning LLM, \textbf{o3-mini}~\citep{openai2025_o3mini_systemcard}. As shown in Table~\ref{tab.stronger_agent}, \ourframework outperforms the considered baselines on average, indicating that the benefits of \ourframework generalize well across model strengths. For weaker LLMs, we conduct experiments with \textbf{GPT-OSS-20B}~\citep{openai2025gptoss120bgptoss20bmodel}, \textbf{Qwen2.5-7B}~\citep{qwen2025qwen25technicalreport}, and \textbf{Qwen2.5-Coder-3B}~\citep{hui2024qwen2_5_coder}, \textbf{GPT-4.1-nano}~\citep{openai2025_gpt41_nano}. We find that these models are unable to reliably follow instructions and often produce syntactically incorrect Python code, suggesting that the meta-agent role requires sufficiently strong capabilities to handle its multiple responsibilities. 


\begin{wraptable}{r}{0.4\textwidth}
\vspace{-1\baselineskip}
\centering
\setlength{\tabcolsep}{2pt}
\resizebox{0.4\columnwidth}{!}{
\begin{tabular}{llccc}
\toprule
\textbf{Agent} & \textbf{Meta-agent} & \multicolumn{1}{c}{\textbf{AIME24}} & \multicolumn{1}{c}{\textbf{GPQA}} & \multicolumn{1}{c}{\textbf{Avg.}} \\
\midrule
GPT-4o & GPT-4o & 33.33 & 50.60 & 41.97 \\
o3-mini & o3-mini & 90.00 & 76.77 & 83.38 \\
\midrule
GPT-4o & o3-mini & 36.67 & 60.10 & 48.38 \\
o3-mini & GPT-4o & 83.33 & 73.74 & 78.54 \\
\bottomrule
\end{tabular}
}
 \caption{\small \ourframework with different models.}
\vspace{-1\baselineskip}
\label{tab.asymmetric_ablation}
\end{wraptable}

\noindent\textbf{Heterogeneous agents.} The previous experiments use the same LLM for both the meta-agent and the individual agents and already achieved strong results. An intriguing question is whether heterogeneous assignments can yield additional benefits or drawbacks. Specifically, we explore pairing a stronger LLM as the meta-agent with a weaker LLM as the individual agent, and vice versa. As shown in Table~\ref{tab.asymmetric_ablation},
when GPT-4o is the individual agent and the meta-agent is replaced with o3-mini, performance improves notably but still falls short of directly using o3-mini for both roles. Conversely, when o3-mini is the individual agent and the meta-agent is replaced with GPT-4o, performance decreases, though it remains better than the setting where GPT-4o is the agent and o3-mini is the meta-agent. These results suggest that while a stronger meta-agent can provide benefits, the overall performance is ultimately constrained by the capability of the individual agent.



\begin{wraptable}{r}{0.5\textwidth}
\vspace{-1\baselineskip}
\centering
\setlength{\tabcolsep}{2pt}
\resizebox{0.5\columnwidth}{!}{
\begin{tabular}{llll}
    \toprule
    \textbf{LLM} & \multicolumn{3}{c}{\textbf{GPT-4o}} \\
    \textbf{Methods} & \textbf{AIME24} & \textbf{GPQA} & \multicolumn{1}{c}{\textbf{Avg.}} \\
    \toprule
    \textbf{\ourframework} & \textbf{33.33} & \textbf{50.60} & \textbf{41.97} \\
    \midrule
    \textbf{\quad - MAS-Init} &  12.50 & 48.43 & 30.46\downred{11.50}\\
    \midrule
    \textbf{\quad - MAS-Evolve} & 20.00 & 48.73 & 34.37\downred{7.60} \\
    \textbf{\quad\quad - meta-design} & 20.83 & 45.18 & 33.01\downred{8.96} \\
    \textbf{\quad\quad - meta-feedback} & 25.00 & 42.17 & 33.59\downred{8.38} \\
    \textbf{\quad\quad → ensemble meta-feedback} & 16.67 & 46.88 & 31.77\downred{10.19} \\
    \midrule
    \textbf{\quad - MAS-Verify} & 6.70 & 33.83 & 20.27\downred{21.70}\\
        \bottomrule
\end{tabular}
}
 \caption{\small Ablations on the three steps in \ourframework.}
\vspace{-1\baselineskip}
\label{tab.ablation}
\end{wraptable}

\noindent\textbf{MAS-Init.} Table~\ref{tab.reasoning_results} suggests that building blocks can achieve strong performance in some problems. To quantify their contribution, we ablate MAS-Init by skipping execution of the building blocks in the first step, letting MAS-Verify judge \textit{solely} based on the five candidate solutions produced by the five iterations of MAS-Evolve. As shown in Table~\ref{tab.ablation}\textbf{(-MAS-Init)}, this significantly degrades performance, highlighting the importance of including MAS-Init and the ability of \ourframework to dynamically revert to building blocks.


\begin{figure*}[t]
\centering
\includegraphics[width=0.9\textwidth]{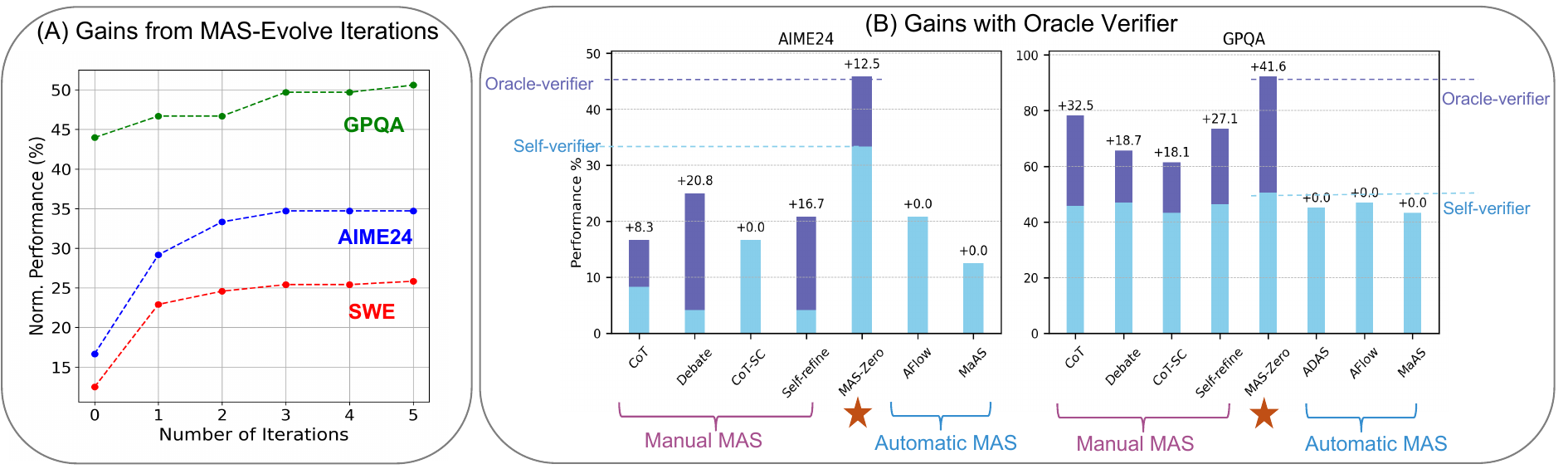}
\vspace{-0.7\baselineskip}
\caption{\small  \textbf{(A) }Performance gains (GPT-4o) over MAS-Evolve. \textbf{(B) } Performance gains (GPT-4o) given an oracle verifier. Automatic MAS baselines
cannot integrate external verifiers, yielding zero improvement.
}
\label{fig.upper_bound_iter}
\vspace{-1.2\baselineskip}
\end{figure*}

\noindent\textbf{MAS-Evolve.} To evaluate its importance, we first conduct an ablation by skipping the entire MAS-Evolve and letting MAS-Verify judge \textit{solely} based on the four building block outputs from MAS-Init. As shown in Table~\ref{tab.ablation}\textbf{(-MAS-Evolve)}, the performance drops notably, indicating that MAS-Evolve is useful for the overall improvements. For \textbf{meta-design}, we modify the prompt to ask the meta-agent to propose a MAS configuration \textit{without} attempting to decompose the question into sub-tasks; Table~\ref{tab.ablation}\textbf{(-meta-design)} shows that removing decomposition leads to a significant performance drop, demonstrating that breaking down the task is a meaningful contributor to the effectiveness of \ourframework. For \textbf{meta-feedback}, we test two variants: (1) modifying the prompt so that the meta-agent critiques the current MAS \textit{without} analyzing the solvability and completeness of each sub-task or LLM agent \textbf{(-meta-feedback)}; (2) since meta-feedback can be noisy due to the self-evolving nature of the system, we explore a straightforward method to improve reliability via ensembling \textbf{(→ ensemble meta-feedback)}. Following \citep{du2023improvingfactualityreasoning}, we generate multiple feedback candidates (three in our experiments) from the meta-agent and then use an additional call to the meta-agent to select the best one. The corresponding rows in Table~\ref{tab.ablation} reveal that removing meta-feedback results in a large performance drop, confirming that it is critical to the overall effectiveness of \ourframework. Surprisingly, the ensemble approach not only fails to improve performance but even reduces it. This counter-intuitive result suggests that the current straightforward meta-feedback is already strong, and advancing \ourframework will require designing more principled strategies for reliable feedback.


\noindent\textbf{Gains from MAS-Evolve at each iteration.} To further evaluate the self-evolving capability in MAS-Evolve, we examine performance across iterations. As shown in Fig.~\ref{fig.upper_bound_iter}(A), accuracy at iteration 0 (before MAS-Evolve, only MAS-Init) and 1 (after the first refinement) is notably lower, indicating that the system struggles to design effective MAS at the outset. With subsequent iterations, however, \ourframework progressively improves, demonstrating a strong ability to self-evolve through the refinement cycle of meta-design, meta-feedback, and the accumulated experience library.

\noindent\textbf{MAS-Verify.} This final step determines which candidate solution is selected as the final answer. To assess its importance, we conducted an ablation where the system simply used the \textit{last} iteration as the output, without any additional judgment. The last row in Table~\ref{tab.ablation} shows a sharp performance decline (the largest drop among all ablations). This is intuitive because, as shown earlier, the ability to dynamically revert to building blocks is indispensable. Yet it is also revealing, since the self-evolving nature of \ourframework might suggest that the final iteration should yield the strongest solution. Instead, the outcome highlights that iterative refinement alone is insufficient, and that effective verification is essential to counteract the stochasticity of pure self-evolving methods without ground-truth signals.


\noindent\textbf{Potential of MAS-Verify with oracle verifier.} We showed that MAS-Verify is crucial, and this highlights substantial headroom for further improvement. Unlike existing automatic MAS frameworks, \ourframework can seamlessly incorporate \textit{external verifiers}, making it naturally positioned to benefit from advances in verification techniques. Fig.~\ref{fig.upper_bound_iter}(B) illustrates this potential: when equipped with an \textit{oracle verifier} that labels outputs as “correct’’ or “incorrect’’ using ground-truth answers, \ourframework’s performance improves dramatically, further widening the gap over both manual and automatic MAS and pushing GPQA close to 95\%. This demonstrates that stronger verification could unlock significant headroom for \ourframework.







\section{Conclusion}


We presented \ourframework, the first \textit{inference-time-only} automatic MAS design framework with \textit{zero} supervision. It iteratively designs and refines MAS, decomposes complex questions, reverts to simpler strategies when sufficient, and verifies candidate answers. Comprehensive experiments show its strong effectiveness, cost-efficiency, and the contribution of each step.
\bibliography{custom}
\bibliographystyle{plain}

\appendix

\newpage
\counterwithin{table}{section}
\renewcommand{\thetable}{\Alph{section}.\arabic{table}}
\counterwithin{figure}{section}
\renewcommand{\thefigure}{\Alph{section}.\arabic{figure}}

\section{\ourframework Algorithm}
\label{ap.algo}

In Section~\ref{sec.framework}, we details the three steps of \ourframework. Algorithm~\ref{alg} presents the detailed algorithm. Highlighting indicates \MyPurple{MAS-Init}, \MyBlue{MAS-Evolve} and \MyOrange{MAS-Verify}.

\begin{algorithm*}
\scriptsize
\caption{\ourframework: Designing Multi-Agent Systems with Zero Supervision}
\label{alg}
\KwIn{Question $Q$, building blocks $\{\mathcal{M}^{(1)}, \ldots, \mathcal{M}^{(k)}\}$, meta-agent $\mathcal{A}$, iterations $T$}
\KwOut{Final Answer $y^*$}

\textbf{Initialize} candidate answers $\mathcal{H}\!\gets[\,]$, experience library $\mathcal{E}\!\gets\varnothing$\;

\stepbox{mypurple}{Step 1: MAS-Init}
\ForEach{building block $\mathcal{M}^{(i)}$}{
  $Y_0^{(i)} \gets \texttt{Execute}(\mathcal{M}^{(i)}, Q)$ \tcp*{Run each building block}
  extract final answer $y_0^{(i)}$ from $Y_0^{(i)}$\;
  append $y_0^{(i)}$ to $\mathcal{H}$\;
  $\mathcal{E} \gets \mathcal{E} \cup \{(Q,\mathcal{M}^{(i)},y_0^{(i)})\}$ \tcp*{Store the answers from MAS-Init}
}


\stepbox{myblue}{Step 2: MAS-Evolve}
    $(\mathcal{Q}_0, \mathcal{M}_0) \gets \mathcal{A}.\texttt{Meta\_Design}\big(
       Q, \{\mathcal{M}^{(i)}\}, \mathcal{E}, \text{Constraints}=\{\mathcal{M}^{(i)}\}\big)$\;
    \tcp{Decompose into sub-tasks $\mathcal{Q}_0$ and assign sub-MAS $\mathcal{M}_0$ grounded in building blocks}
    \For{$t=1$ \KwTo $T$}{
      $Y_t \gets \texttt{Execute}(\mathcal{M}_{t-1}, \mathcal{Q}_{t-1})$ \tcp*{Run current MAS on sub-tasks}
      extract sub-task outputs $\{(x^{\text{sub}}_j,y^{\text{sub}}_j)\}$ and agent outputs $\{(x^{\text{agent}}_\ell,y^{\text{agent}}_\ell)\}$ from $Y_t$\;
    
      $(\mathcal{Q}_t,\mathcal{M}_t,y_t,f_t) \gets \mathcal{A}.\texttt{Meta\_Feedback}\big(
         Q,\mathcal{Q}_{t-1},\mathcal{M}_{t-1},
         \{(x^{\text{sub}}_j,y^{\text{sub}}_j)\},\{(x^{\text{agent}}_\ell,y^{\text{agent}}_\ell)\},
         \mathcal{E},\text{Constraints}=\{\mathcal{M}^{(i)}\}\big)$\;
      \tcp{Assess solvability and completeness; revise decomposition and sub-MAS}
    
      $\mathcal{E} \gets \mathcal{E} \cup \{(\mathcal{Q}_{t-1},\mathcal{M}_{t-1},Y_t,f_t)\}$ \tcp*{Store sub-tasks, sub-MAS, intermediate outputs, and feedback}
    
      \If{$y_t \neq \bot$}{append $y_t$ to $\mathcal{H}$}
    }
\stepbox{myorange}{Step 3: MAS-Verify}
$y^* \gets \mathcal{A}.\texttt{Self\_Verify}(\mathcal{H})$ \tcp*{Select final answer from all candidates}

\Return $y^*$\;
\end{algorithm*}

\section{Implementations, Benchmarks and Baselines Details}
\label{ap.benchmark_baseline}

\paragraph{Implementation details.} 

As described in Sec.~\ref{sec.experiments}, \ourmodel produces 9 candidate answers. For fair comparison, we sample 9 independent outputs for CoT-SC and take the majority vote. Similarly, both debate and self-refine are run for {9} rounds. All models are accessed through their respective APIs.\footnote{{\color{black}We use TogetherAI API (\url{https://www.together.ai/}) for Llama and Qwen.}} 
Temperature for meta-agent is set to 0.5. For baselines, we strictly use parameters found in original papers and provided code.


\paragraph{Benchmarks}
Table~\ref{tab.data_size} shows the detailed statistics for each dataset. For BrowseComp and Frames, we randomly sample 10\% for testing, due to the large dataset size. We evaluate SWE using its official code available at \url{https://github.com/SWE-bench/SWE-bench/}.

\begin{table}[h]
\centering
\setlength{\tabcolsep}{2pt}
\resizebox{0.6\columnwidth}{!}{
\begin{tabular}{llllll}
\toprule
\textbf{Split} & \multicolumn{1}{l}{\textbf{AIME24}} & \multicolumn{1}{l}{\textbf{GPQA}} & \multicolumn{1}{l}{\textbf{SWE}} & {\textbf{BrowseComp}} & {\textbf{Frames}}\\
\midrule
\textbf{Validation} & 6 & 32 & 60 & --- & ---\\
\textbf{Test} & 24 & 166 & 240 & 126 & 82\\
\bottomrule
\end{tabular}
}
\vspace{-1\baselineskip}
\caption{\small Data size for each split in each dataset.}
\label{tab.data_size}
\end{table}

\paragraph{Baselines details} We use the official implementations of all baselines, sourced directly from their public repositories. For manual MAS methods, this includes ReConcile (\url{https://github.com/dinobby/ReConcile}), MAD (\url{https://github.com/Skytliang/Multi-Agents-Debate}), SPP (\url{https://github.com/MikeWangWZHL/Solo-Performance-Prompting}), and DyLAN (\url{https://github.com/SALT-NLP/DyLAN}). For automatic MAS methods, this includes ADAS (\url{https://github.com/ShengranHu/ADAS}), AFlow (\url{https://github.com/FoundationAgents/MetaGPT/tree/main/examples/aflow}), and MaAS (\url{https://github.com/bingreeky/MaAS}).

\section{Cost Computation}
\label{ap.cost}

In Fig.~\ref{fig.pareto_front}, we report the cost of single-agent systems, manual MAS, automatic MAS, and \ourframework. Costs are computed using OpenAI’s official pricing as of May 2025 at
\url{https://openai.com/api/pricing/}. To ensure accuracy, we track usage directly via the official OpenAI API field \texttt{``response.usage''} for all methods. As a result, the reported values reflect the actual monetary cost, accounting for both input and output tokens.

\section{Additional Related Work}
\label{ap.related_work}



In Sec.~\ref{sec.related_work}, we briefly introduced the most important related works and highlighted their contrast with \ourframework. In this section, we provide a more detailed discussion of existing works. For completeness, we also note that a number of training-based approaches have been proposed, but we omit them from Sec.~\ref{sec.related_work} since \ourframework does not involve updating LLM parameters.

Some prior work treats prompt  optimization for individual agents as part of MAS design. Examples include PromptBreeder~\citep{fernando2024promptbreeder}, DsPy~\citep{khattab2023dspycompilingdeclarativelanguage} and Self-Discover \citep{zhou2024selfdiscover}. 
More recently, this idea has been extended to broader automatic MAS design, where prompt optimization is included either as an additional design step or as part of the search space.

\paragraph{Manual MAS design.} In addition to the approaches discussed in Sec.~\ref{sec.related_work}, several other methods fall into this family. DyLAN~\citep{liu2024dynamicllmpowered} uses message passing to dynamically activate agent compositions; Reconcile~\citep{chen-etal-2024-reconcile} and MAD~\citep{liang2023encouraging} employ debate and round-table discussion, SPP~\citep{wang2023unleashing} leverages collaboration among multiple personas.

\paragraph{Automatic MAS design.} We follow the categories introduced in Sec.~\ref{sec.related_work} and additionally include the training-based family.

\noindent\textbf{Val-Pruning.} This line of work has evolved quickly~\citep{zhang2024cutcrapeconomicalcommunication,zhang2025gdesignerarchitectingmultiagentcommunication,hu2024selfevolvingmultiagentcollaborationnetworks}.
Earlier examples include GPTSwarm~\citep{zhuge2024languageagentsoptimizablegraphs} which optimizes graph structures via reinforcement learning but struggles to represent workflows with conditional state due to limitations of static graphs.  AgentSquare~\citep{shang2024agentsquare} leverages a verifier as a performance predictor to guide the pruning.

\noindent\textbf{Val-Generation.} Besides the approaches introduced in Sec.~\ref{sec.related_work} that employ a meta-agent to generate building block connections and configurations, another line of work uses the meta-agent to directly generate the required agents or blocks. For example, AutoAgents~\citep{chen2024autoagentsframeworkautomaticagent} and AgentVerse~\citep{chen2023agentversefacilitatingmultiagentcollaboration} create specialized agents through a planner agent, while EvoAgent~\citep{yuan2024evoagent} applies evolutionary algorithms to optimize this generation process. Similarly, Symbolic-MoE~\citep{chen2025symbolicmixtureofexpertsadaptiveskillbased} leverages validation signals to construct block profiles and select the best-performing planner agent.

\noindent\textbf{Training-based.} More recent work attempts to explicitly train the agents or meta-agents within MAS. For example, ReMA~\citep{wan2025remalearningmetathinkllms} and OWL~\citep{hu2025owloptimizedworkforcelearning} train the agents in a manual MAS, while MAS-GPT~\citep{ye2025masgpttrainingllmsbuild} collects data from off-the-shelf MAS and trains a meta-agent via supervised fine-tuning (SFT). FlowReasoner~\citep{gao2025flowreasonerreinforcingquerylevelmetaagents} builds on this by extending SFT with reinforcement learning (RL). Puppeteer~\citep{dang2025multiagentcollaborationevolvingorchestration} directly RL trains the meta-agent in an end-to-end manner.




\section{Standard deviation for the Experiments}
\label{ap.error_bar}

To confirm the statistical significance of the experimental results in Table~\ref{tab.reasoning_results}, we repeat the experiment \textit{three} times, following \citep{zhang2024aflowautomatingagenticworkflow,liu2024dynamicllmpowered}. We can see that the MAS can exhibit high variance due to the inherent nature of multi-agent systems: the variance may be amplified by the interactions among multiple agents~\citep{cemri2025multiagentllmsystemsfail}, and the generated temperature of the agents is typically non-zero. Nevertheless, \ourframework is significantly stronger than other baselines, with only one exception in AIME24 and two in GPQA, as mentioned in Sec.~\ref{sec.experiments}.

\begin{table*}[t]
\centering
\setlength{\tabcolsep}{2pt}

\begin{minipage}{0.48\textwidth}
\centering
\resizebox{0.7\columnwidth}{!}{
\begin{tabular}{llll}
\toprule
\textbf{LLMs} & \textbf{GPT-4o} & \textbf{Llama3.3} & \textbf{Qwen2.5} \\
\midrule
CoT & $\pm${1.97} & $\pm${1.96} & $\pm${0.00} \\
CoT-SC & $\pm${3.40} & $\pm${5.20} & $\pm${1.96} \\
Debate & $\pm${7.08} & $\pm${7.08} & $\pm${5.20} \\
Self-Refine & $\pm${3.93} & $\pm${1.97} & $\pm${1.97} \\
ReConcile & $\pm${1.97} & $\pm${1.96} & $\pm${1.97} \\
MAD & $\pm${1.96} & $\pm${3.40} & $\pm${3.40} \\
SPP & $\pm${5.20} & $\pm${1.96} & $\pm${5.89} \\
DyLAN & $\pm${1.96} & $\pm${0.00} & $\pm${3.93} \\
ADAS & $\times$ & $\pm${5.30} & $\pm${6.38} \\
AFlow & $\pm${1.96} & $\pm${1.96} & $\pm${0.00} \\
MAS-GPT & $\pm${3.54} & $\pm${1.96} & $\pm${5.20} \\
\textbf{\ourframework} & $\pm${5.89} & $\pm${3.15} & $\pm${5.20} \\
\bottomrule
\end{tabular}
}
\caption{\small Standard deviations for AIME24.}
\label{tab.std_dev_aime24}
\end{minipage}
\hfill
\begin{minipage}{0.48\textwidth}
\centering
\resizebox{0.7\columnwidth}{!}{
\begin{tabular}{llll}
\toprule
\textbf{LLMs} & \textbf{GPT-4o} & \textbf{Llama3.3} & \textbf{Qwen2.5} \\
\midrule
CoT & $\pm${1.29} & $\pm${0.89} & $\pm${2.72} \\
CoT-SC & $\pm${1.36} & $\pm${0.78} & $\pm${1.94} \\
Debate & $\pm${0.41} & $\pm${0.78} & $\pm${1.07} \\
Self-Refine & $\pm${2.08} & $\pm${2.08} & $\pm${3.01} \\
ReConcile & $\pm${1.29} & $\pm${2.43} & $\pm${1.85} \\
MAD & $\pm${1.28} & $\pm${1.28} & $\pm${3.08} \\
SPP & $\pm${1.74} & $\pm${1.25} & $\pm${1.20} \\
DyLAN & $\pm${3.44} & $\pm${0.49} & $\pm${0.75} \\
ADAS & $\pm${3.83} & $\pm${3.83} & $\pm${3.98} \\
AFlow & $\pm${1.70} & $\pm${2.48} & $\pm${1.77} \\
MAS-GPT & $\pm${1.02} & $\pm${1.02} & $\pm${2.25} \\
\textbf{\ourframework} & $\pm${1.67} & $\pm${0.51} & $\pm${2.08} \\
\bottomrule
\end{tabular}
}
\caption{\small Standard deviations for GPQA.}
\label{tab.std_dev_gpqa}
\end{minipage}

\end{table*}

\section{Illustration of MAS-Evolve}
\label{ap.meta_iter}

\begin{figure*}[h]
\centering
\includegraphics[width=0.8\textwidth]{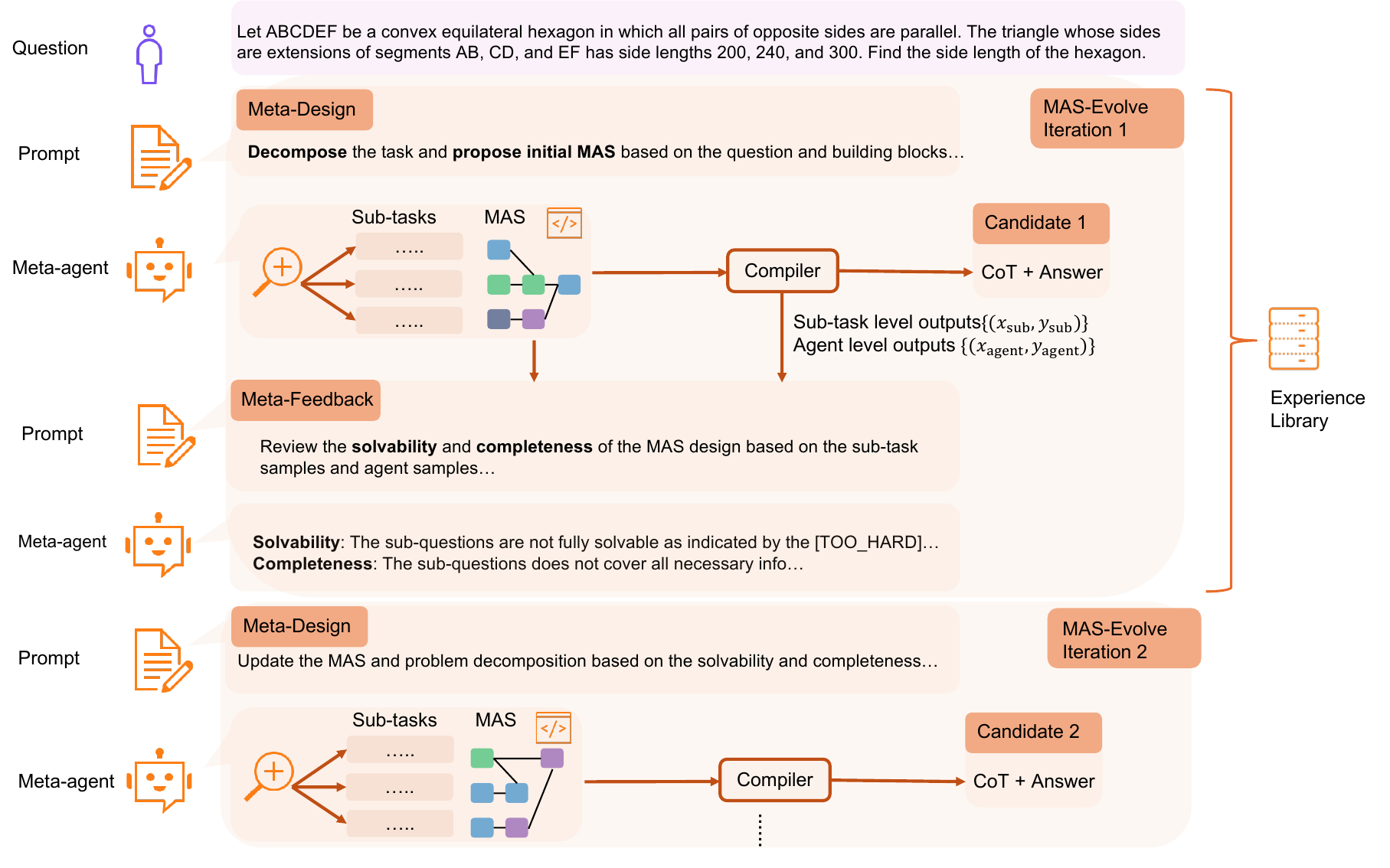}
\vspace{-1\baselineskip}
\caption{\small Illustration for the iterations in MAS-Evolve. 
}
\label{fig.smas_detail}
\vspace{-5mm}
\end{figure*}

Fig.~\ref{fig.smas_detail} illustrates MAS-Evolve. Given a question and the building blocks, the meta-agent is prompted to decompose the task and propose a MAS (see Appendix~\ref{ap.prompt_detail} for detailed prompts). The meta-agent then generates a MAS in the form of code, which is executed by an external compiler to produce intermediate and final outputs for the sub-tasks and agents.

After this meta-design and execution, the meta-feedback phase begins. In this phase, both the resulting MAS and its intermediate outputs are provided to the meta-agent to review their \textit{solvability} and \textit{completeness}. Based on this evaluation, the meta-agent generates targeted feedback. The MAS, its intermediate outputs, and the feedback are stored in the \textit{experience library}, which is then used as additional context to refine the design in subsequent iterations.

\section{Example of MAS produced from \ourframework}
\label{ap.vis}

\begin{figure*}[t]
\centering
\includegraphics[width=0.9\textwidth]{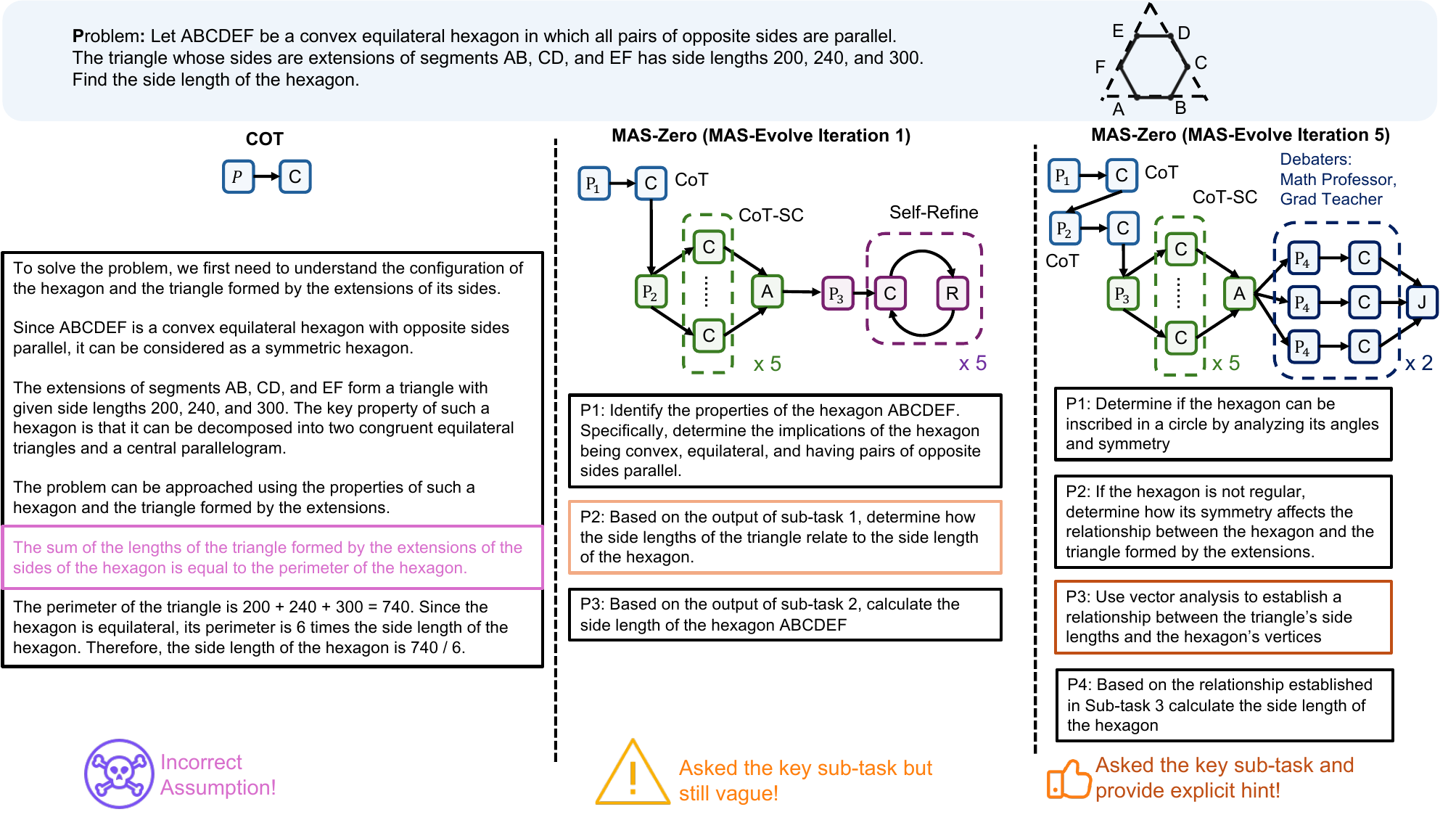}
\caption{\small 
An example illustrates how the MAS produced by \ourframework outperforms both CoT and early iterations. In this case, \ourframework learns to decompose the task into 4 sub-tasks and dynamically assign appropriate sub-MAS: CoT for the first two, CoT-SC (sampling 5 completions) for the third, and Debate (2 rounds with a math professor and a graduate teacher as debaters) for the fourth. 
}
\vspace{-1\baselineskip}
\label{fig.mas_moment}
\end{figure*}

\ourframework learns to decompose a new question and assign appropriate sub-MAS to each sub-task dynamically.
This type of dynamic assignment would have been difficult to design manually. Fig.~\ref{fig.mas_moment} showcases the effectiveness of \ourframework by demonstrating how it can construct and refine MAS architectures on the fly, adapting complexity to the requirements of the task.

\section{Prompt Details}
\label{ap.prompt_detail}

As described in Sec.~\ref{sec.framework}, \ourframework uses prompts to implement its three steps. In this section, we present all detailed prompts used, including the building blocks (Fig.~\ref{fig.cot}, \ref{fig.cot_sc}, \ref{fig.debate}, \ref{fig.self_refine}) in MAS-Init; the Meta-Design (Fig.~\ref{fig.prompt_meta_design}) and Meta-Feedback (Fig.~\ref{fig.prompt_meta_feedback}) in MAS-Evolve; and the prompts for MAS-Verify (Fig.~\ref{fig.verifier_prompt}).


\begin{figure*}[ht]
\centering
\begin{tcolorbox}[colback=outerboxcolor!20,colframe=innerboxcolor!75,title=Prompt for \texttt{Meta-Design},fonttitle=\bfseries,  fontupper=\small,arc=3mm,boxrule=1pt]
\textbf{Overview.}
You are an expert machine learning researcher testing various agentic systems. You are given a set of building blocks (e.g., CoT, CoT-SC, Self-Refine, Debate) and a question. Each building block is represented as executable code and can contain one or more LLM agents with specialized settings (instruction, temperature, etc.). A \textit{sub-MAS} refers to one or more building blocks assigned together to solve a specific sub-task. The overall MAS is formed by coordinating multiple sub-MAS to solve the full problem.

\textbf{Your objectives are:}
\begin{itemize}[leftmargin=*,noitemsep,topsep=3pt]
\item \textbf{Perform task decomposition.} Decompose the given question into \textit{manageable yet interdependent} sub-tasks, such that each sub-task is specific and detailed enough for a sub-MAS (formed from building blocks) to solve. Do not solve the tasks yourself or leak the expected answer. Instead, design the decomposition so that the sub-tasks are easier for sub-MAS to solve, and justify how they combine to yield the final answer.

\textbf{Make sure}
\begin{itemize}[leftmargin=*,noitemsep,topsep=3pt]
\item Each sub-task should explicitly build on the outputs of prior sub-tasks.
\item The final sub-task should naturally yield the overall answer to the original question.
\end{itemize}

\item \textbf{Design sub-MAS assignments.} Based on the resulting sub-tasks, assign one or more building blocks to form a sub-MAS for each sub-task. You may adjust block parameters (e.g., temperature, number of debate rounds), but you must not invent new blocks or prune the provided ones without grounding.

\textbf{For example:} Given available building blocks \{CoT, CoT-SC, Self-Refine, Debate\}, and the resulting sub-tasks: sub-task 1, sub-task 2, sub-task 3:

    \begin{itemize}[leftmargin=*,noitemsep,topsep=3pt]
    
    \item \textbf{Step 1: For each sub-task, specify its sub-MAS.} 
    \begin{itemize}[leftmargin=*,noitemsep,topsep=3pt]
        \item It may use a \textbf{single block} (e.g., Sub-task 1: CoT).
        
        \item  It may use a \textbf{sequential chain of blocks} (e.g., Sub-task 3: CoT $\rightarrow$ Self-Refine).  
        
        \item  Or it may use \textbf{parallel blocks inside the sub-task} (e.g., Sub-task 2: \{CoT $\parallel$ Debate\}, meaning both blocks process the same input).  
    \end{itemize}

    \item \textbf{Step 2: Connect the sub-tasks (sub-MAS).}     After defining sub-MAS, specify how the sub-tasks depend on one another: 
    \begin{itemize}[leftmargin=*,noitemsep,topsep=3pt]

        \item \textbf{Sequential connection}: Sub-tasks are connected in a linear chain, where the output of one becomes the input to the next.\\
        Example: [CoT] (sub-task 1) $\rightarrow$ [Debate] (sub-task 2) $\rightarrow$ [CoT $\rightarrow$ Self-Refine] (sub-task 3).

        \item \textbf{Parallel connection}: Multiple sub-tasks depend on the same earlier sub-task’s output and run in parallel.\\
        Example: [CoT] (sub-task 1) $\rightarrow$ \{[Debate] (sub-task 2), [CoT $\rightarrow$ Self-Refine] (sub-task 3)\}. \\
        Here, both sub-task 2 and sub-task 3 consume the result of sub-task 1 in parallel.

    \end{itemize}

    \textbf{IMPORTANT:} Do not collapse all decomposed sub-tasks into a single instruction handled by one block. Each sub-task must be addressed by its own sub-MAS.
    \end{itemize}

\end{itemize}
\textbf{Final remark:} Your aim is to design an optimal block connection that can perform well on each sub-task. Your code should implement the existing blocks as-is. Do not propose new blocks or modify existing ones; you may only adjust their connections and settings (e.g., instruction, temperature).

\end{tcolorbox}
\vspace{-1\baselineskip}
\caption{Prompt for \texttt{Meta-Design}. Additional examples, building blocks code and output format instruction are omitted for clarity.}
\label{fig.prompt_meta_design}
\end{figure*}

\begin{figure*}[ht]
\centering
\begin{tcolorbox}[colback=outerboxcolor!20,colframe=innerboxcolor!75,title=Prompt for \texttt{Meta-Feedback},fontupper=\small,fonttitle=\bfseries,arc=3mm,boxrule=1pt]

\textbf{Overview.}  
You are given a candidate Multi-Agent System (MAS) design, including: (i) its executable code, (ii) the sub-task outputs from each sub-MAS, (iii) the outputs of individual agents, (iv) the final response, and (v) experience of prior iterations. Your task is to critically evaluate this MAS and provide feedback to guide refinement in the next iteration.  

\begin{itemize}[leftmargin=*,noitemsep,topsep=3pt]
\item \textbf{Solvability}: 
Assess whether all sub-tasks are solvable by the corresponding block via checking the output answer of each sub-task. 
\begin{itemize}[leftmargin=*,noitemsep,topsep=3pt]
\item If the output explicitly includes [TOO\_HARD], this means the sub-task is too difficult and should be further decomposed.  
\item If the output is incorrect, identify whether the issue is due to 
    \begin{itemize}[leftmargin=*,noitemsep,topsep=3pt]
    
    \item Insufficient decomposition (the task is still too hard) or 
    \item Some agents in the block are malfunctioning, or the underlying LLM is too weak to solve the sub-task. This can be checked by reviewing the agent outputs:  
    (a) If the agent itself is not optimal (e.g., poor instruction, temperature, etc.), the settings need to be improved.  
    (b) If the agent architecture is not optimal, a new block should be proposed by recombining existing blocks or adjusting their settings.  
    
\end{itemize}

    Please \textbf{justify} whether it is (i), the decomposition issue or (ii) the block and agent issue. It could also be both. When proposing new sub-task, \textbf{make sure} 
        \begin{itemize}[leftmargin=*,noitemsep,topsep=3pt]
        \item It is specific and detailed enough to solve and to contribute to the final answer;  
        \item All required information is carried over from previous sub-tasks or provided in the instruction;  
        \item The outputs from related sub-tasks are correctly incorporated (e.g., added to the \texttt{taskInfo} list when calling the agent);  
        \item The connection to prior sub-tasks is explicit (e.g., instructions should state “Based on the output of sub-task $i$”).  
        \end{itemize}
\end{itemize}

\item \textbf{Completeness:} Do the sub-tasks include all necessary information from the original query that can ensure the aggregation of sub-task responses can effectively yield a comprehensive answer to the user query? Note that while a sub-task might include only part of the necessary information, it is not allowable for any particular piece of critical information to be omitted from all sub-tasks. 

    \begin{itemize}[leftmargin=*,noitemsep,topsep=3pt]
        \item If critical information is missing, refine the decomposition to include it.  
        \item Ensure sub-tasks are connected so that the aggregated outputs can yield a correct and comprehensive final answer.  
    \end{itemize}
\end{itemize}

Now, you need to improve or revise the implementation, or implement the new proposed MAS based on the reflection.

\end{tcolorbox}
\vspace{-1\baselineskip}
\caption{Prompt for \texttt{Meta-Feedback}. Additional examples, building blocks code and output format instruction are omitted for clarity.}
\vspace{-1\baselineskip}
\label{fig.prompt_meta_feedback}
\end{figure*}

\begin{figure*}[ht]
\centering
\begin{tcolorbox}[colback=outerboxcolor!20,colframe=innerboxcolor!75,title=Prompt for \texttt{MAS-Verify},fontupper=\small,fonttitle=\bfseries,arc=3mm,boxrule=1pt]
Given the problem and a list of candidate answers, carefully review the reasoning steps and final answers to select the most reliable candidate. Do not solve the task yourself. 

In your output, use the \texttt{``thinking''} field to compare the selected answer with each unselected answer one by one, identify the erroneous steps in the unselected answers, and give a detailed explanation of why they are incorrect.  
In the \texttt{``selection''} field, output the ID of the best answer.  

\medskip
Problem: \{problem\}  

Answer List: \{candidate answers\}
\end{tcolorbox}
\vspace{-1\baselineskip}
\caption{Prompt for \texttt{MAS-Verify}. The \texttt{``candidate answers''} have already been ranked and filtered before being passed to the meta-agent. Additional examples and output format instructions are omitted for clarity.}
\label{fig.verifier_prompt}
\end{figure*}

\section{Code Template}
\label{ap.code_template}

\ourframework uses a code template to aid MAS code generation to filling in a specific \texttt{forward} function. Fig.~\ref{fig.util_code} shows how the utility code is provided. Fig.s~\ref{fig.cot}, \ref{fig.cot_sc}, \ref{fig.debate}, and \ref{fig.self_refine} show the implementations of each building blocks.

\section{Usage of Large Language Models}
We use LLMs solely for grammar polishing.

\begin{figure*}[ht]
\centering
\begin{tcolorbox}[colback=outerboxcolor!20,colframe=innerboxcolor!75,title=Utility Code,fonttitle=\bfseries,arc=3mm,boxrule=1pt]
\begin{lstlisting}[style=mypython]
# Named tuple for holding task information
Info = namedtuple('Info', ['name', 'author', 'content', 'prompt', 'sub_tasks', 'agents', 'iteration_idx'])

# Format instructions for LLM response
FORMAT_INST = lambda request_keys: f"Reply EXACTLY with the following JSON format.\n{str(request_keys)}\nDO NOT MISS ANY FIELDS AND MAKE SURE THE JSON FORMAT IS CORRECT!\n"

# Description of the role for the LLM
ROLE_DESC = lambda role: f"You are a {role}."

class LLMAgentBase():

    def __init__(self, output_fields: list, agent_name: str,
                 role='helpful assistant', model=None, temperature=None) -> None:
        self.output_fields = output_fields
        self.agent_name = agent_name

        self.role = role
        self.model = model
        self.temperature = temperature
        # give each instance a unique id
        self.id = random_id()
        

    def generate_prompt(self, input_infos, instruction) -> str:
        # generate prompt based on the input_infos
        # ...

    def query(self, input_infos: list, instruction, iteration_idx=-1) -> dict:
        # call generate_prompt and the LLM to get output
        # ...

    def __repr__(self):
        return f"{self.agent_name} {self.id}"



class AgentArchitecture:
    """
    Fill in your code here.

    def forward(self, taskInfo) -> Union[Info, str]:
        Args:
        - taskInfo (Info): Task information.
        
        Returns:
        - Answer (Info): Your FINAL Answer.
    """
\end{lstlisting}
\end{tcolorbox}
\vspace{-1\baselineskip}
\caption{Utility code. }
\label{fig.util_code}
\end{figure*}

\begin{figure*}[ht]
\centering
\begin{tcolorbox}[colback=outerboxcolor!20,colframe=innerboxcolor!75,title=Implementation of CoT Building Blocks,fonttitle=\bfseries,arc=3mm,boxrule=1pt]
\begin{lstlisting}[style=mypython]
def forward(self, taskInfo):
    # Instruction for the Chain-of-Thought (CoT) approach
    # It is an important practice that allows the LLM to think step by step before solving the task.
    cot_instruction = self.cot_instruction

    # Instantiate a new LLM agent specifically for CoT
    # To allow the LLM to think before answering, we need to set an additional output field 'thinking'.
    cot_agent = LLMAgentBase(['thinking', 'answer'], 'Chain-of-Thought Agent',  model=self.node_model, temperature=0.0)

    # Prepare the inputs for the CoT agent
    # The input should be a list of Info, and the first one is often the taskInfo
    cot_agent_inputs = [taskInfo]

    # Get the response from the CoT agent
    thinking, answer = cot_agent(cot_agent_inputs, cot_instruction)
    final_answer = self.make_final_answer(thinking, answer)
    
    # Return only the final answer
    return final_answer   
\end{lstlisting}
\end{tcolorbox}
\vspace{-1\baselineskip}
\caption{Implementation of CoT building blocks}
\label{fig.cot}
\end{figure*}

\begin{figure*}[ht]
\centering
\begin{tcolorbox}[colback=outerboxcolor!20,colframe=innerboxcolor!75,title=Implementation of CoT-SC Building Blocks ,fonttitle=\bfseries,arc=3mm,boxrule=1pt]
\begin{lstlisting}[style=mypython]

def forward(self, taskInfo):
    # Instruction for step-by-step reasoning
    cot_instruction = self.cot_instruction
    N = self.max_sc # Number of CoT agents

    # Initialize multiple CoT agents with a higher temperature for varied reasoning
    cot_agents = [LLMAgentBase(['thinking', 'answer'], 'Chain-of-Thought Agent',  model=self.node_model, temperature=0.5) for _ in range(N)]

    # Majority voting function to select the most common answer
    from collections import Counter
    def majority_voting(answers):
        return Counter(answers).most_common(1)[0][0]
    
    thinking_mapping = {}
    answer_mapping = {}
    possible_answers = []
    for i in range(N):
        thinking, answer = cot_agents[i]([taskInfo], cot_instruction)
        possible_answers.append(answer.content)
        thinking_mapping[answer.content] = thinking
        answer_mapping[answer.content] = answer

    # Ensembling the answers from multiple CoT agents
    answer = majority_voting(possible_answers)

    thinking = thinking_mapping[answer]
    answer = answer_mapping[answer]

    final_answer = self.make_final_answer(thinking, answer)

    return final_answer  
\end{lstlisting}
\end{tcolorbox}
\vspace{-1\baselineskip}
\caption{Implementation of CoT-SC building blocks}
\label{fig.cot_sc}
\end{figure*}

\begin{figure*}[ht]
\centering
\begin{tcolorbox}[colback=outerboxcolor!20,colframe=innerboxcolor!75,title=Implementation of Debate Building Blocks,fonttitle=\bfseries,arc=3mm,boxrule=1pt]
\begin{lstlisting}[style=mypython]

def forward(self, taskInfo):
    # Instruction for initial reasoning
    debate_initial_instruction = self.cot_instruction

    # Instruction for debating and updating the solution based on other agents' solutions
    debate_instruction = "Given solutions to the problem from other agents, consider their opinions as additional advice. Please think carefully and provide an updated answer. Put your thinking process in the 'thinking' field and the updated answer in the 'answer' field. "
    
    # Initialize debate agents with different roles and a moderate temperature for varied reasoning
    debate_agents = [LLMAgentBase(['thinking', 'answer'], 'Debate Agent',  model=self.node_model, role=role, temperature=0.5) for role in self.debate_role]

    # Instruction for final decision-making based on all debates and solutions
    final_decision_instruction = "Given all the above thinking and answers, reason over them carefully and provide a final answer. Put your thinking process in the 'thinking' field and the final answer in the 'answer' field."
    final_decision_agent = LLMAgentBase(['thinking', 'answer'], 'Final Decision Agent',  model=self.node_model, temperature=0.0)

    max_round = self.max_round # Maximum number of debate rounds
    all_thinking = [[] for _ in range(max_round)]
    all_answer = [[] for _ in range(max_round)]

    # Perform debate rounds
    for r in range(max_round):
        for i in range(len(debate_agents)):
            if r == 0:
                thinking, answer = debate_agents[i]([taskInfo], debate_initial_instruction)
            else:
                input_infos = [taskInfo] + [all_thinking[r-1][i]] + all_thinking[r-1][:i] + all_thinking[r-1][i+1:]
                thinking, answer = debate_agents[i](input_infos, debate_instruction)
            all_thinking[r].append(thinking)
            all_answer[r].append(answer)
    
    # Make the final decision based on all debate results and solutions
    thinking, answer = final_decision_agent([taskInfo] + all_thinking[max_round-1] + all_answer[max_round-1], final_decision_instruction)
    final_answer = self.make_final_answer(thinking, answer)

    return final_answer
\end{lstlisting}
\end{tcolorbox}
\vspace{-1\baselineskip}
\caption{Implementation of Debate building blocks}
\label{fig.debate}
\end{figure*}

\begin{figure*}[ht]
\centering
\begin{tcolorbox}[colback=outerboxcolor!20,colframe=innerboxcolor!75,title=Implementation of Self-Refine Building Blocks,fonttitle=\bfseries,arc=3mm,boxrule=1pt]
\begin{lstlisting}[style=mypython]
def forward(self, taskInfo):
    # Instruction for initial reasoning
    cot_initial_instruction = self.cot_instruction

    # Instruction for reflecting on previous attempts and feedback to improve
    cot_reflect_instruction = "Given previous attempts and feedback, carefully consider where you could go wrong in your latest attempt. Using insights from previous attempts, try to solve the task better."
    cot_agent = LLMAgentBase(['thinking', 'answer'], 'Chain-of-Thought Agent', model=self.node_model, temperature=0.0)

    # Instruction for providing feedback and correcting the answer
    critic_instruction = "Please review the answer above and criticize on where might be wrong. If you are absolutely sure it is correct, output exactly 'True' in 'correct'."
    critic_agent = LLMAgentBase(['feedback', 'correct'], 'Critic Agent', model=self.node_model, temperature=0.0)
    
    N_max = self.max_round # Maximum number of attempts
    
    # Initial attempt
    cot_inputs = [taskInfo]
    thinking, answer = cot_agent(cot_inputs, cot_initial_instruction, 0)

    for i in range(N_max):
        # Get feedback and correct status from the critic
        feedback, correct = critic_agent([taskInfo, thinking, answer], critic_instruction, i)
        if correct.content == 'True':
            break
            
        # Add feedback to the inputs for the next iteration
        cot_inputs.extend([thinking, answer, feedback])

        # Reflect on previous attempts and refine the answer
        thinking, answer = cot_agent(cot_inputs, cot_reflect_instruction, i + 1)

    final_answer = self.make_final_answer(thinking, answer)

    return final_answer

\end{lstlisting}
\end{tcolorbox}
\vspace{-1\baselineskip}
\caption{Implementation of Self-Refine building blocks}
\label{fig.self_refine}
\end{figure*}

\end{document}